%% file: main.tex
\documentclass[10pt,twocolumn,letterpaper]{article}

\usepackage[pagenumbers]{cvpr} 


\definecolor{cvprblue}{rgb}{0.21,0.49,0.74}
\usepackage[pagebackref,breaklinks,colorlinks,allcolors=cvprblue]{hyperref}

\usepackage[accsupp]{axessibility}  

\usepackage{graphicx,subcaption} 
\usepackage{booktabs,multirow,array} 
\usepackage[table]{xcolor} 
\usepackage[labelfont=bf]{caption} 
\usepackage[utf8]{inputenc} 

\usepackage[table]{xcolor} 
\definecolor{red4}{HTML}{febf92}
\definecolor{blue4}{HTML}{77b1e2}
\definecolor{green4}{HTML}{F5FFFA}
\usepackage{xcolor} 

\newcommand{\ourmethod}{Scone}
\newcommand{\ourmethodFullname}{\underline{\textbf{S}}ubject \underline{\textbf{co}}mposition and distinctio\underline{\textbf{n}} \underline{\textbf{e}}nhancement}
\newcommand{\ourbench}{SconeEval}
\newcommand{\qwenvl}{Qwen3-VL-30B-A3B-Instruct}
\newcommand{\qwen}{Qwen3-30B-A3B-Instruct-2507}
\newcommand{\qwenedit}{Qwen-Image-Edit-2509}
\newcommand{\flux}{Flux.1-dev}

\usepackage{pifont}   
\usepackage{xspace}   
\newcommand{\icoyes}{\textcolor{ForestGreen}{\ding{51}}\xspace} 
\newcommand{\icono}{\textcolor{Red}{\ding{55}}\xspace}          

\def\paperID{}
\def\confName{CVPR}
\def\confYear{2026}

\title{\includegraphics[height=1.1em]{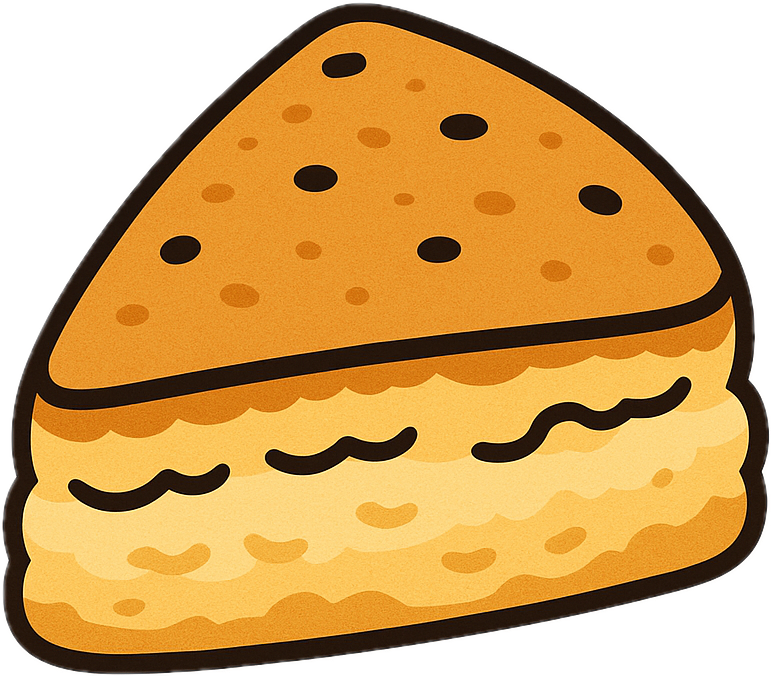}\hspace{4.5pt} Scone: Bridging Composition and Distinction in Subject-Driven Image Generation via Unified Understanding-Generation Modeling}

\author{
    Yuran Wang\textsuperscript{1,2}\thanks{Equal contribution}
    \quad~Bohan Zeng\textsuperscript{1,2}\footnotemark[1]
    \quad~Chengzhuo Tong\textsuperscript{1,2}
    \quad~Wenxuan Liu\textsuperscript{1}
    \quad~Yang Shi\textsuperscript{1,2} \\
    Xiaochen Ma\textsuperscript{4}
    \quad~Hao Liang\textsuperscript{1}
    \quad~Yuanxing Zhang\textsuperscript{2}
    \quad~Wentao Zhang\textsuperscript{1,3,5}\thanks{Corresponding author: wentao.zhang@pku.edu.cn} \\      
    \textsuperscript{1}Peking University
    \quad~\textsuperscript{2}Kling Team, Kuaishou Technology 
    \quad~\textsuperscript{3}Zhongguancun Academy \\
    \quad~\textsuperscript{4}HKUST 
    \quad~\textsuperscript{5}Beijing Key Laboratory of Data Intelligence and Security (Peking University)
}

\begin{document}
\maketitle

\begin{abstract}

Subject-driven image generation has advanced from single- to multi-subject composition, while neglecting distinction, the ability to distinguish and generate the correct subject when inputs contain multiple candidates. This limitation restricts effectiveness in complex, realistic visual settings. We propose Scone, a unified understanding-generation method that integrates composition and distinction. Scone enables the understanding expert to act as a semantic bridge, conveying semantic information and guiding the generation expert to preserve subject identity while minimizing interference. A two-stage training scheme first learns composition, then enhances distinction through semantic alignment and attention-based masking. We also introduce SconeEval, a benchmark for evaluating both composition and distinction across diverse scenarios. Experiments demonstrate that Scone outperforms existing open-source models in composition and distinction tasks on two benchmarks. Our model, benchmark, and training data are available at: https://github.com/Ryann-Ran/Scone.

\end{abstract}

\begin{figure}[t]
    \centering
    \includegraphics[width=0.95\columnwidth]{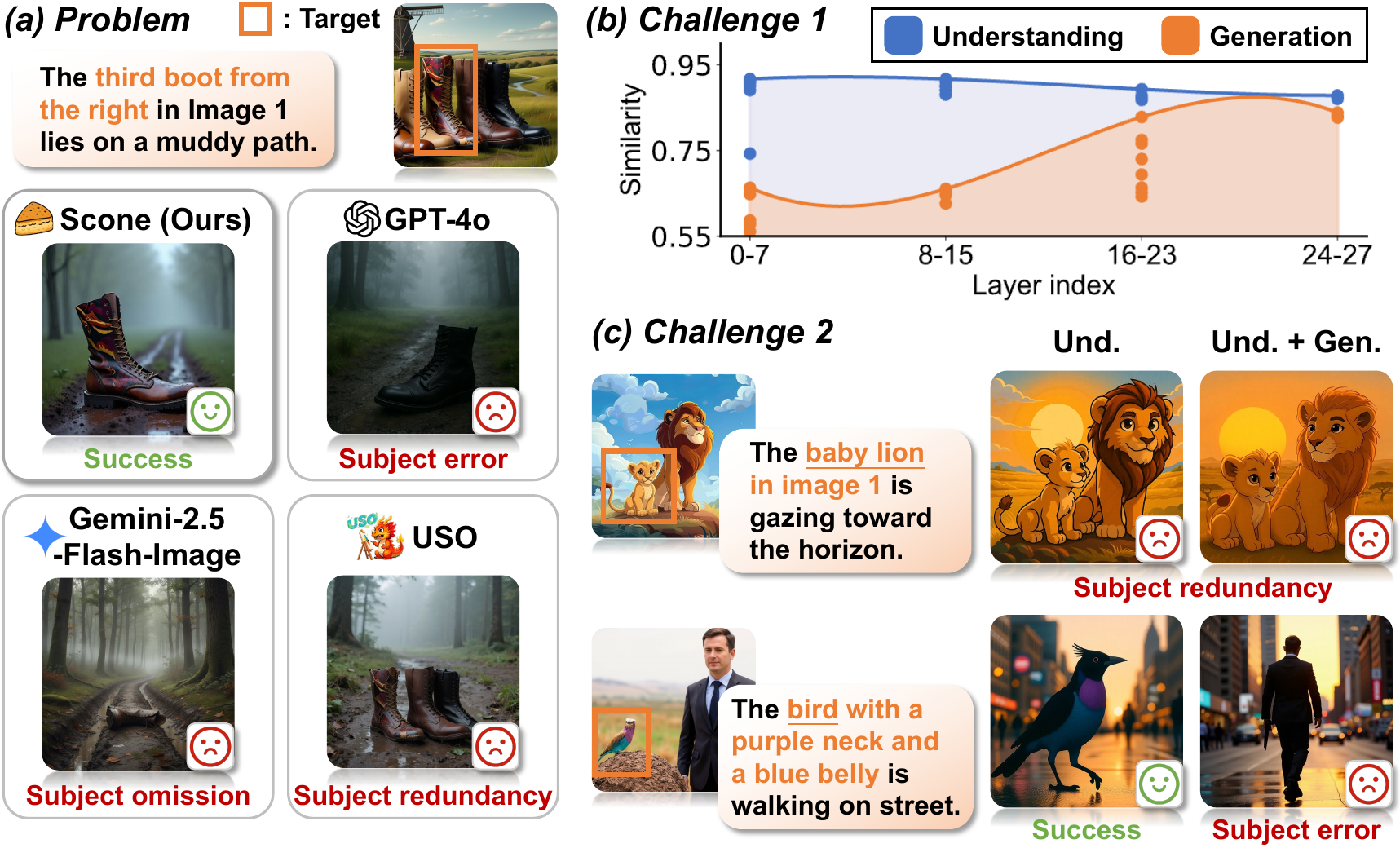}
    \caption{\textbf{The distinction problem and challenges.}
    \textbf{(a) Problem.}
    State-of-the-art methods have limitations in distinguishing target subjects specified by the instruction.
    \textbf{(b) Challenge 1: semantic deficiency in generation.} Reference image information from the understanding and generation experts in the unified model is used to compute semantic similarity with instruction.
    \textbf{(c) Challenge 2: biased understanding and misaligned generation.}
    ``Und.'' and ``Und.+Gen.'' indicate whether textual information from  generation expert in the unified model is included to collaborate with understanding expert.
    The unified model is BAGEL~\cite{deng2025bagel}.}
    \label{fig:problem}
\end{figure}

\section{Introduction}

Image generation methods~\cite{google2025gemini, wu2025qwenimagetechnicalreport,ge2025expand} have demonstrated exceptional capabilities, enabling the generation of desired images across diverse scenarios~\cite{wang2024devil}.
Subject-driven image generation has recently gained significant attention, with the focus evolving from single-subject to multi-subject generation, incorporating more input images. Existing methods~\cite{xiao2025omnigen, wu2025less, wu2025omnigen2, wu2025qwenimagetechnicalreport} can process two or more input images and combine subjects based on instructions.
Moreover, methods such as~\cite{google2025gemini, ye2025echo4oharnessingpowergpt4o} extend this capability by accepting more than four images, showcasing potential for more complex composition tasks.

However, existing works primarily focus on expanding subject combinations while neglecting the ability to distinguish target subjects in complex contexts. As shown in~\cref{fig:problem}(a), although current models can combine multiple subjects, they may fail to distinguish and generate the correct target subject when a reference image contains multiple candidates, leading to problems such as subject omissions (none of the candidate subjects appear) or errors (misidentification of the target subject).
Real-world images often involve interference and intricate details~\cite{10638128,liu2025motion}, further limiting practical performance.
Thus, we emphasize examining the input subjects themselves, focusing on the model’s ability to \emph{\textbf{distinguish the target subject within complex contexts and leverage this information for generation}}.

A core challenge is extracting useful information from complex references, which remains difficult for generation models. Subject distinction relies on \textbf{semantic understanding} of instruction’s expression of references, where understanding models are more proficient~\cite{lin2025perceive,an2024mc,zhang2025cfbench}.
As shown in~\cref{fig:problem}(b), in a unified understanding-generation model consisting of an understanding expert and a generation expert, the information encoded by the understanding expert is more similar to the instruction, which means more aligned with instruction than that encoded by the generation expert. This reveals generation models' deficiency and understanding model's advantage in interacting with instructions and semantically understanding reference information.
%
%
However, this semantic advantage of understanding models is not entirely reliable: understanding models often exhibit biases~\cite{tang2025video, zhang2024holmes, lei2023revealing,liu2025sota}, which become problematic when directly used to assist generation.
As illustrated in~\cref{fig:problem}(c), in a unified model, relying only on semantic information from understanding expert still struggles to prevent irrelevant subjects from appearing, and subject errors persist even with correct semantic information due to misalignment between generation and understanding experts.

\begin{figure}[t]
    \centering
    \includegraphics[width=0.95\columnwidth]{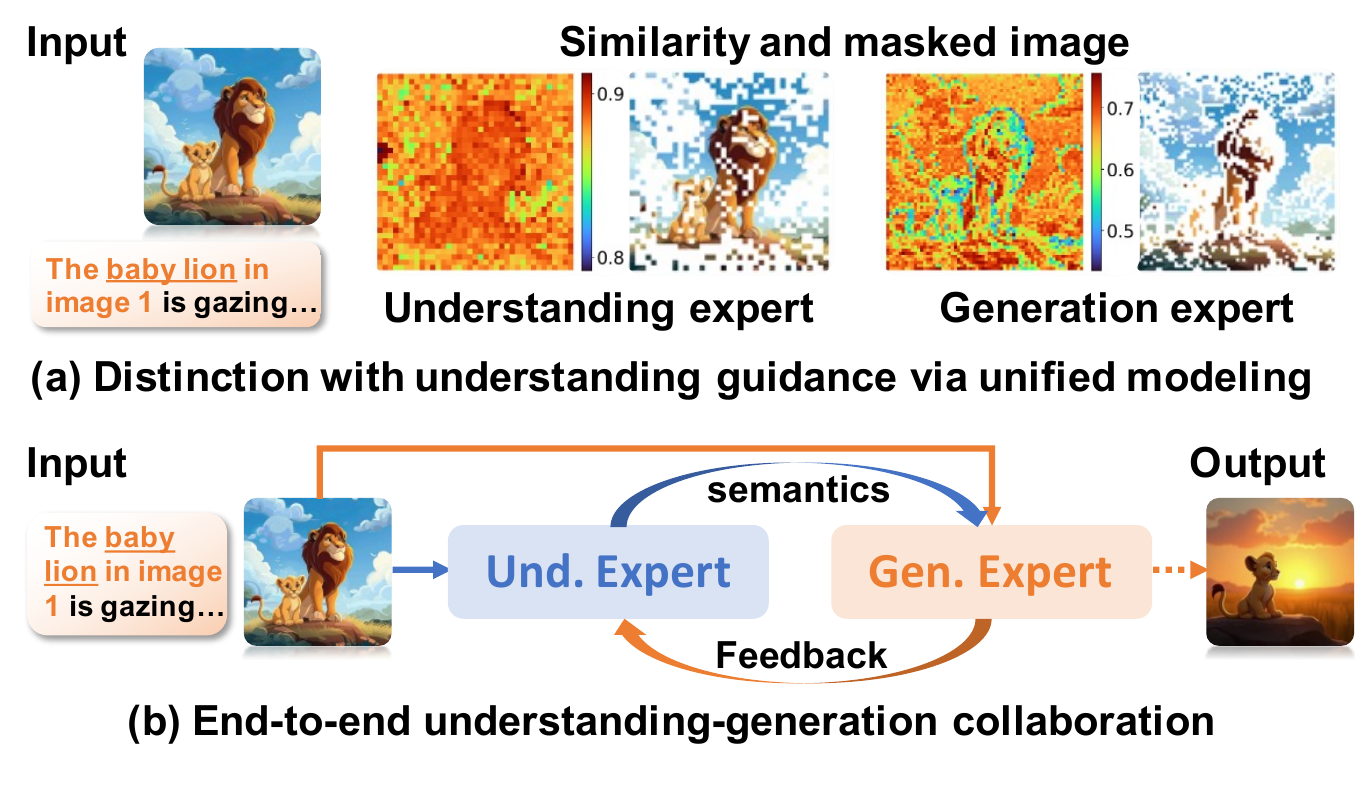}
    \caption{\textbf{Our motivation.} (a) visualizes the early similarity between image token  from the understanding and generation experts and text token within the unified model, showing that the former attends to semantic regions while the latter is less sensitive.
    (b) illustrates the collaboration between the understanding and generation experts within the unified model through end-to-end training.}
    \label{fig:motivation}
\end{figure}

Compared with generation models, unified models offer a clear advantage for subject-driven image generation because the understanding expert captures semantic cues earlier than the generation expert~\cite{Zhang_2025_CVPR}, as illustrated in~\cref{fig:motivation}(a). These early-layer semantics highlight instruction-relevant regions such as candidate subjects and enable more accurate distinction in complex reference images.
%
Moreover, to alleviate bias introduced by the understanding expert, the unified architecture allows end-to-end collaboration, as shown in~\cref{fig:motivation}(b). The understanding expert refines its semantic interpretation through feedback from generation, and the generation expert aligns with these cues to better preserve subject-related details.

Based on these insights, we propose a subject-driven image generation method, \textbf{\ourmethod}~(\ourmethodFullname), built upon a unified understanding-generation model capable of handling subject composition and distinction.
Our method leverages the strong understanding capabilities of the understanding expert to overcome the limitations of the generation expert in complex contexts involving reference images and instructions.
Specifically, \ourmethod~enables the understanding expert to act as a \textbf{\emph{semantic bridge}} conveying high-level semantic information to guide generation, which called \textbf{understanding bridge strategy}.
In the first training stage, the model learns subject composition on single-candidate data (\ie a reference image contains only one candidate subject).
In the second stage, the understanding expert is trained to align visual and textual representations and filter instruction-irrelevant regions using a semantic mask derived from early layer, forming a robust semantic bridge.
After this formation, the understanding expert provides semantic guidance to the generation expert, ensuring that subject-related information is emphasized while unrelated interference is suppressed.
This design enables \ourmethod~to distinguish useful reference information and achieve precise subject composition in complex multi-subject contexts through \textbf{\textit{internal}} understanding, without relying on external models, additional parameters, or test-time techniques.
Unlike methods with external understanding modules, our unified architecture jointly learns semantics and generation, avoiding extra latency and preserving end-to-end optimization.
As shown in~\cref{fig:problem}(a), compared to existing methods, our method more accurately distinguishes relevant reference information and generates ideal results.

Furthermore, to evaluate whether existing models can genuinely distinguish subjects in reference images based on instructions and use relevant information to generate the correct target subject, we introduce a new benchmark, \textbf{\ourbench}, which includes subject-driven image generation tasks with varying difficulty levels, including composition, distinction, and distinction \& composition. This benchmark provides a comprehensive evaluation from both composition and distinction perspectives.


Our main contributions are threefold:
\begin{itemize}
    \item We propose the \textbf{\ourmethod}~model, which supports multi-subject composition and excels in subject distinction in complex contexts, ranking first among open-source models on OmniContext benchmark.
    \item We introduce the \textbf{understanding bridge strategy}, which transforms the understanding expert into a semantic bridge, enabling early multimodal alignment and attention-based semantic masking to guide the generation expert, enhancing subject distinction and semantic fidelity without adding extra parameters.
    \item We develop \textbf{\ourbench}, a challenging benchmark to evaluate subject-driven image generation from both composition and distinction perspectives.
\end{itemize}

\section{Related work}

\subsection{Subject-driven image generation}
Early subject-driven generation rely on fine-tuned diffusion models~\cite{ye2023ip, wang2024instantid, zeng2024ipdreamer} with image conditions for flexible customization. With the rise of Diffusion Transformer~\cite{peebles2023scalable}, generation quality improves significantly.
Recent methods~\cite{labs2025flux1kontextflowmatching, kim2024instantfamily, tan2025ominicontrol, xiao2025omnigen} extend single-subject to multi-subject composition, but they typically assume clean references and are prone to interference from irrelevant information under complex conditions, causing deviations from the target subject.
Although methods like SSR-Encoder~\cite{zhang2024ssr} aim to isolate features, they handle only simple prompts with a single reference, reflecting limited understanding and restricted effectiveness under complex instructions or noisy inputs.

\subsection{Unified understanding-generation models}
To advance general-purpose agents, several methods~\cite{chen2025janus, xie2024show, xie2025show, deng2025bagel, lin2025uniworld, chen2025opengpt, li2025uniworld,song2025dualtoken} integrate multimodal understanding and generation tasks within a unified architecture. By leveraging multimodal understanding, these methods enhance the stability of image generation when handling complex instructions. Some methods~\cite{wu2025omnigen2, ye2025echo4oharnessingpowergpt4o,an2025unictokens} use this capability for subject-driven generation. However, when reference images contain substantial irrelevant content, existing unified models lack effective mechanisms to prevent interference, often resulting in unwanted subjects. We address this gap by using understanding semantics to better distinguish target conditions and guide cleaner, more reliable generation.

\section{The \ourmethod~model}

We present \textbf{\ourmethod}~(\ourmethodFullname), which supports multi-subject composition and demonstrates strong distinction capability in complex contexts via unified understanding-generation modeling.

\subsection{Motivation and preliminaries}
\label{sec:motivation}

\paragraph{Distinction with understanding guidance via unified modeling}
Unified models outperform generation models in complex semantics due to stronger understanding ability and better text-image alignment~\cite{tang2025exploring}.
The understanding expert captures instruction-relevant semantics earlier than the generation expert, before textual features emerge~\cite{Zhang_2025_CVPR,tang2025exploring}.
As shown in~\cref{fig:motivation}(a), early-layer similarity with text token indicates that the understanding expert attends to key subject regions, while the generation expert is less semantically sensitive.

\begin{figure}[t]
    \centering
    \includegraphics[width=0.95\columnwidth]{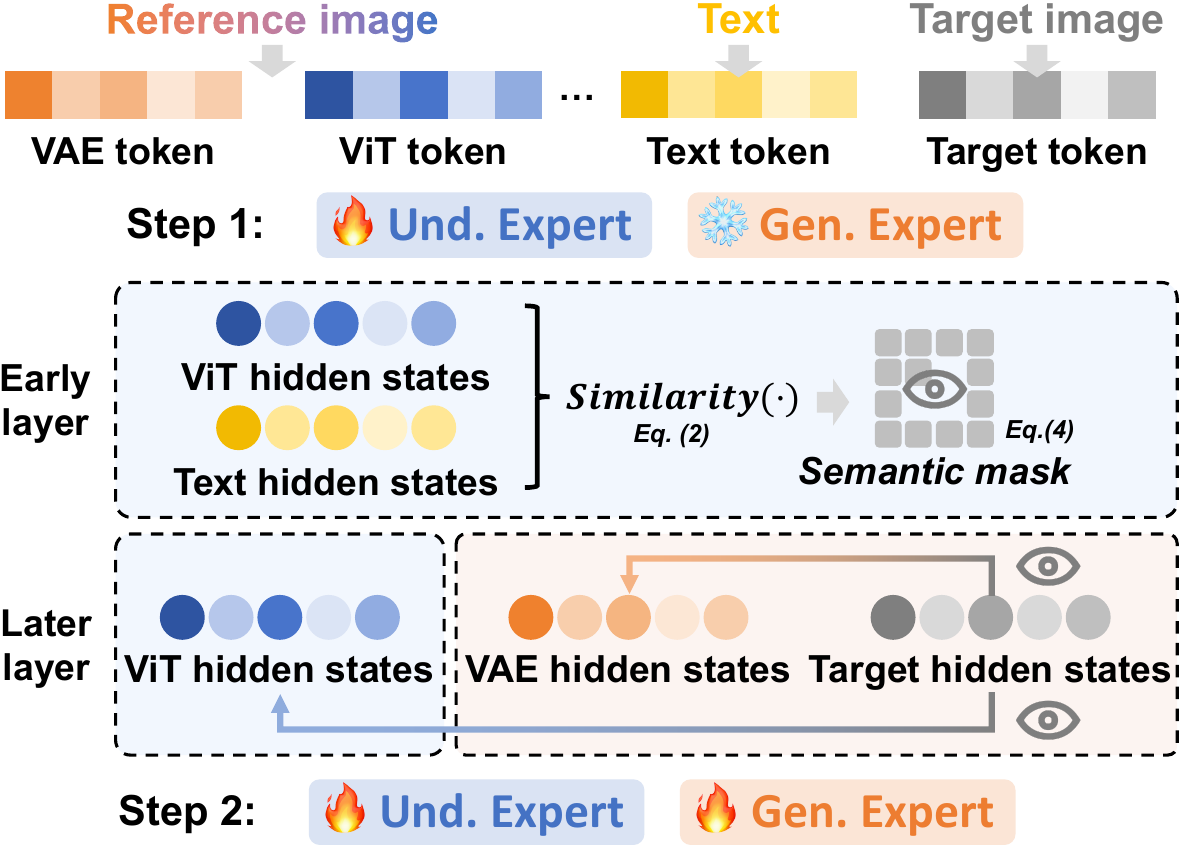}
    \caption{\textbf{Understanding bridge strategy.} \textbf{Step 1: Understanding bridge formation.} Early semantic alignment and attention masking enable the understanding expert to serve as the \textit{semantic bridge}.
    \textbf{Step 2: Understanding bridge guidance.} The generation expert is optimized under the guidance of the semantic bridge, enabled by unified understanding-generation modeling.}
    \label{fig:framework}
\end{figure}

\paragraph{End-to-end understanding-generation collaboration}
The understanding expert may introduce semantic bias, leading to subject errors or redundancy. Unified modeling enables end-to-end collaboration, as shown in~\cref{fig:motivation}(b). The understanding expert refines its semantics through generation feedback, and the generation expert aligns with these cues to preserve subject-related details in complex reference images.

\subsection{Unified understanding-generation modeling}
We adopt BAGEL~\cite{deng2025bagel} as base, a Mixture-of-Transformer-Experts architecture that handles understanding and generation information through dedicated experts sharing multimodal attention. We optimize it using the original MSE loss, with no additional parameters. For subject-driven generation, image tokens from Vision Transformer (ViT) encoder and instruction tokens are handled by the understanding expert, while image tokens from the VAE model are processed by the generation expert.
To improve distinction in complex contexts, the understanding expert acts as a semantic bridge that provides discriminative cues for generation.

\subsection{Stage \uppercase\expandafter{\romannumeral1}: Composition training}
We first finetune BAGEL on single-candidate data, where each reference image contains a single subject. The understanding expert and generation expert (including corresponding MLP connectors) are trained, while the ViT and VAE remain frozen. One epoch of base data enables both single- and multi-subject generation. A refined dataset is then used for another epoch to further enhance subject consistency. Training data details are provided in~\cref{sec:training_data}.

\subsection{Stage \uppercase\expandafter{\romannumeral2}: Distinction training with understanding bridge strategy}

We propose the \textbf{\emph{understanding bridge strategy}}, which enables the understanding expert to act as a \textbf{\emph{semantic bridge}} that transfers high-level semantic information for generation guidance, as shown in~\cref{fig:framework}. It comprises two steps: forming the semantic bridge via multimodal alignment and guiding generation through this bridge. This design improves subject identity preservation, relevance discrimination, and contextual fidelity.
Multi-candidate data are introduced for distinction-aware composition.

\paragraph{Step 1: Understanding bridge formation}
The understanding expert jointly learns visual and textual semantics to become the semantic bridge. Let $\mathbf{h}^{v} = \{\mathbf{h}^{v}_i\}_{i=1}^{N_v}$ and $\mathbf{h}^{t} = \{\mathbf{h}^{t}_j\}_{j=1}^{N_t}$ denote early-layer \textit{visual and textual hidden states in understanding expert}, respectively.
We apply L2-normalization and compute the cosine similarities as:
\begin{equation}
\mathbf{S} = \hat{\mathbf{H}}^{v} (\hat{\mathbf{H}}^{t})^{\top}, \quad S_{i,j} = \hat{\mathbf{h}}^{v}_i \cdot \hat{\mathbf{h}}^{t}_j = \frac{\mathbf{h}^{v}_i}{\|\mathbf{h}^{v}_i\|_2} \cdot \frac{\mathbf{h}^{t}_j}{\|\mathbf{h}^{t}_j\|_2}.
\end{equation}
The semantic relevance for each visual token is defined as:
\begin{equation}
s_i = \frac{1}{N_t} \sum_{j=1}^{N_t} S_{i,j}.
\end{equation}

We construct a binary semantic mask $\mathbf{M}$ based on a threshold $\tau$ (with parameter study in~\cref{sec:param}), which influences the number of reference image tokens that remain invisible to the \textit{target image tokens in generation expert}. Rather than discarding tokens, the mask modifies the attention logits. For logits $\mathbf{A}$ mapping target tokens to reference image tokens in subsequent layers, we apply the mask as follows:

\begin{equation}
\tilde{A}_{k,i} = A_{k,i} + M_i, \quad M_i =
\begin{cases}
0, & s_i > \tau,\\
-\infty, & \text{otherwise}.
\end{cases}
\label{eq:mask}
\end{equation}
Tokens where $M_i=-\infty$ receive zero attention, which allows target tokens to disregard irrelevant regions. This mechanism establishes the understanding expert as a semantic bridge to align representations and suppress semantic interference. We train the model for 1k steps.

\paragraph{Step 2: Understanding bridge guidance}
Functioning as the semantic bridge, the understanding expert guides the generation expert. We train both experts for an additional 1k steps to align generation representations with the bridge and focus on key regions identified by the understanding expert. This phase enforces semantic consistency within complex compositional scenarios.

\section{The \ourbench~benchmark}
\subsection{Overview}
Existing benchmarks usually offer simple contexts where the reference image contains a single prominent subject referred to by a basic category term.
Such settings fail to reflect performance in real-world images with substantial interference and less structured compositions.
They also mainly assess subject reproduction and composition using similarity metrics from models such as DINOv2~\cite{oquab2023dinov2} and CLIP~\cite{radford2021learning}. In multi-subject settings, averaging similarity across subjects cannot reliably measure generation quality, especially when subject omission or redundancy occurs.

\begin{figure*}[ht]
\centering
\includegraphics[width=0.98\textwidth]{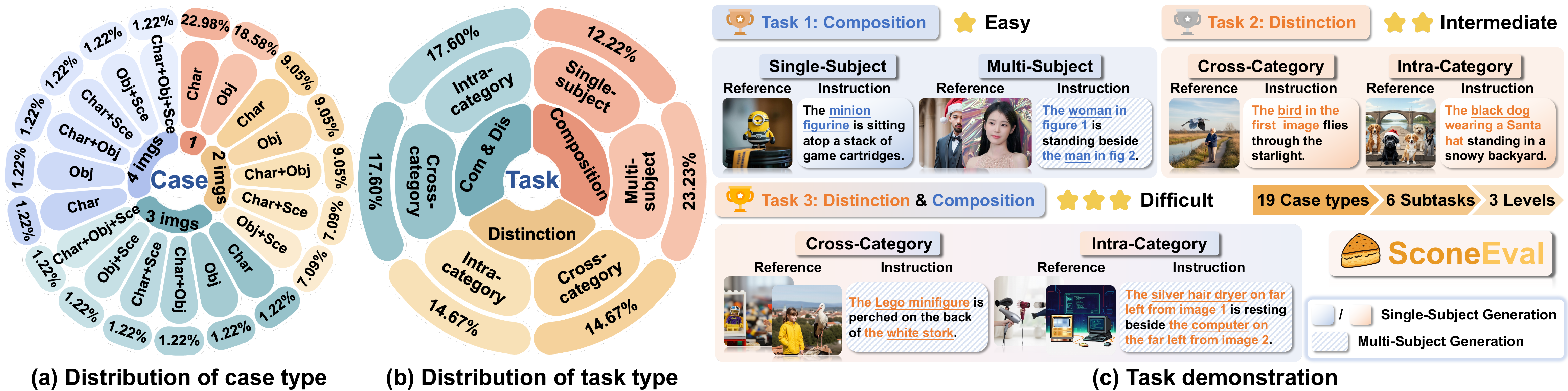}
\caption{
\textbf{Overview of \ourbench.} ``Char'': character, ``Obj'': object, ``Sce'': scene.
\ourbench~evaluates target subject identification and generation in complex contexts, with 409 test cases across three domains, 19 case types, and 6 subtasks, covering composition, distinction, and distinction \& composition tasks.
}
\label{fig:benchmark_task}
\end{figure*}

To evaluate a model’s ability to distinguish and generate referred subjects in complex contexts, we introduce \textbf{\ourbench}, a benchmark with 409 test cases spanning character, object, and scene combinations and subject distinction. It covers 19 case types in~\cref{fig:benchmark_task}(a) and 6 subtasks in~\cref{fig:benchmark_task}(b), enabling comprehensive evaluation of subject distinction and feature utilization.
Unlike traditional benchmarks emphasizing visual fidelity or text alignment, \ourbench~focuses on cross-modal reasoning over complex reference images and instructions, requiring the model to decide \textit{whom} to generate among multiple candidates.
\ourbench~includes three progressively challenging tasks, as shown in~\cref{fig:benchmark_task}(c): composition, distinction, and distinction \& composition. In composition, each reference image contains one subject, and one or more images are used for single- or multi-subject generation. In distinction, each reference image contains multiple subjects, and the model must generate the target one. The distinction \& composition task combines both settings, requiring multi-subject generation from reference images that each contain multiple subjects. Tasks involving distinction include cross-category and intra-category cases, indicating whether candidate subjects in a reference image belong to the same category.
As shown in~\cref{tab:benchmark_comparison}, existing benchmarks mainly evaluate subject composition in simple contexts, while our benchmark targets more realistic scenarios.

\subsection{Construction pipeline}
\label{sec:benchmark_pipeline}

\paragraph{Step 1: Image collection}
Images are collected from three sources.
\textit{(1) Existing benchmarks:} We use \qwenvl~\cite{qwen3technicalreport} to filter images, recognize subjects, and classify subject categories, followed by manual verification to ensure each image contains only one subject. Samples are drawn from DreamBench++~\cite{peng2024dreambench++} and OmniContext~\cite{wu2025omnigen2}.
\textit{(2) T2I (text-to-image) synthesis} and \textit{(3) Open access:} To increase category diversity, we further supplement the benchmark with single-candidate images synthesized by \flux~\cite{flux2024} and additional open-access samples.
Finally, we build a single-candidate image pool covering three categories, character, object, and scene, with 15 subcategories and at least 30 images per subcategory. The images are grouped into sets of 1 to 4 and split into two subsets for subsequent single- and multi-candidate data construction.

\begin{figure}[t]
\centering
\includegraphics[width=0.95\columnwidth]{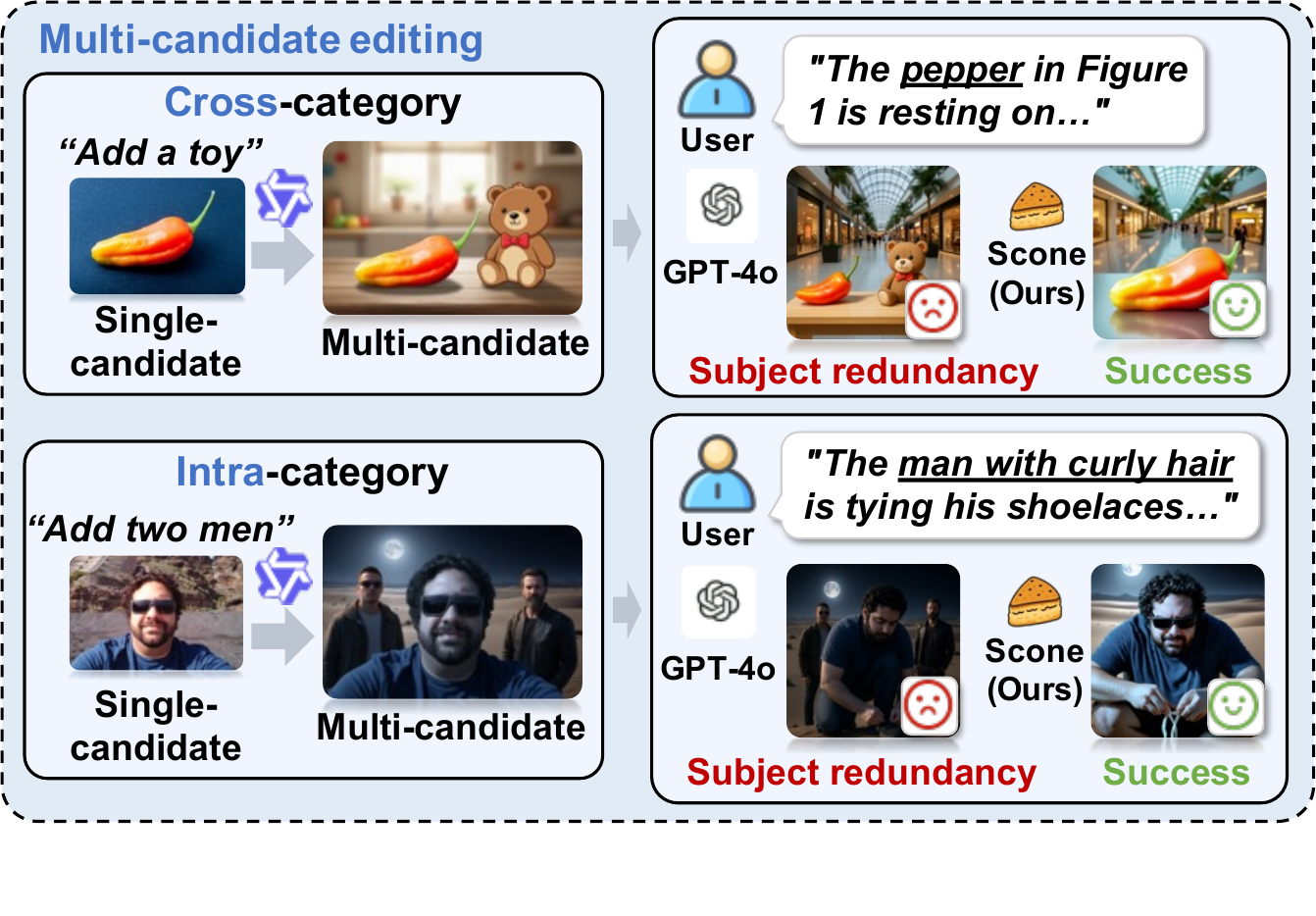}
\caption{
    \textbf{Multi-candidate editing in \ourbench~construction.}
    Task difficulty increases with image and instruction complexity.
}
\label{fig:benchmark_pipeline}
\end{figure}

\paragraph{Step 2: Multi-candidate editing}
We create multi-candidate images by adding other subjects to single-candidate images with \qwenedit~\cite{wu2025qwenimagetechnicalreport}, as shown in~\cref{fig:benchmark_pipeline}, followed by manual verification to ensure each subject is clearly recognizable.

\paragraph{Step 3: Instruction construction}
We adopt a two-step decoupling strategy that separates visual understanding from text generation, reducing cross-image interference and improving subject identification and linguistic coherence.
Instructions explicitly specify target indices or distinct features to avoid ambiguity (\eg ``Image 1'', ``the man with green hair'').
\textit{Step 1: Subject identification (image-to-text).}
Each image is processed independently by the VLM model \qwenvl~\cite{qwen3technicalreport} to identify its most prominent subject, minimizing mutual interference.
For single-candidate images, we extract direct subject names (\eg ``woman'');
for multi-candidate images, we generate referential names with distinctive cues such as attribute, size, and position (\eg ``woman on the left of the image''), guided by the corresponding single-candidate images. For scene images, we provide detailed descriptions to support interaction-related instructions (\eg ``place the bird on the shelf'').
\textit{Step 2: Instruction generation (text-to-text).}
Only the subject names or scene descriptions from Step 1 are fed to the LLM model \qwen~\cite{qwen3technicalreport}, without image inputs.

\subsection{Evaluation protocol}
Following VIEScore~\cite{ku2024viescore} and OmniContext~\cite{wu2025omnigen2}, we use GPT-4.1~\cite{openai2025gptapi} to score composition capability on a 0--10 scale with rationales, covering prompt following and subject consistency. Unlike automatic metrics such as VSM~\cite{eldesokey2025mind} and AlphaCLIP~\cite{sun2024alpha}, which require additional masks, GPT-4.1 provides anchor-free evaluation for complex multi-condition cases.
Prompt for composition scoring is similar to OmniContext~\cite{wu2025omnigen2}, but evaluates subject consistency only for the target subject.
For distinction evaluation, GPT-4.1 judges whether the described reference subject \emph{appears} in the target image, from which we compute accuracy, precision, recall, and F1. Precision and recall reflect redundancy and omission, and the distinction score is defined as the average of accuracy and F1.
To align with the composition scale, we rescale the distinction score from [0,1] to [0,10]. The overall score averages composition and distinction.

\begin{table}[tbp]
  \centering
  \caption{\textbf{Task comparison of existing benchmark for subject-driven image generation.}}
\resizebox{1\columnwidth}{!}{
    \begin{tabular}{cccc}
    \toprule
    \textbf{Benchmark} & \textbf{Composition} & \textbf{Distinction} & \textbf{Distinction \& Composition} \\
    \midrule
    DreamBench~\cite{ruiz2023dreambooth} & \icoyes     & \icono     & \icono \\
    DreamBench++~\cite{peng2024dreambench++} & \icoyes     & \icono     & \icono \\
    OmniContext~\cite{wu2025omnigen2} & \icoyes     & \icono     & \icono \\
    XVerseBench~\cite{chen2025xverse} & \icoyes     & \icono     & \icono \\
    \midrule
    \ourbench~(Ours)  & \icoyes     & \icoyes     & \icoyes \\
    \bottomrule
    \end{tabular}}
  \label{tab:benchmark_comparison}%
\end{table}%

\begin{table*}[t]
  \centering
  \caption{\textbf{Quantitative comparison of existing models on OmniContext~\cite{wu2025omnigen2} benchmark.} ``Char. + Obj.'' indicates Character + Object. ${\dagger}$ indicates our base model.
  Best scores in each group are highlighted in \textbf{bold}.
  }
  \resizebox{1\textwidth}{!}{
    \begin{tabular}{p{12em}ccccccccccc}
    \toprule
    \multirow{2}[4]{*}{\textbf{Method}} & \multicolumn{2}{c}{\textbf{SINGLE}~$\uparrow$} &       & \multicolumn{3}{c}{\textbf{MULTIPLE}~$\uparrow$} &       & \multicolumn{3}{c}{\textbf{SCENE}~$\uparrow$} & \multirow{2}[4]{*}{\textbf{Average~$\uparrow$}} \\
\cmidrule{2-3}\cmidrule{5-7}\cmidrule{9-11}

    \multicolumn{1}{c}{} & \textbf{Character} & \textbf{Object} &       & \textbf{Character} & \textbf{Object} & \textbf{Char. + Obj.} &       & \textbf{Character} & \textbf{Object} & \textbf{Char. + Obj.} &  \\
    \midrule

    \rowcolor{red4!20} \multicolumn{12}{c}{\textit{\textbf{Closed-source model}}} \\
    \midrule
    
    Gemini-2.5-Flash-Image~\cite{google2025gemini} & 8.79      &  \textbf{9.12}     &       &   8.27    & 8.60      &  7.71     &       &  7.63     &  7.65     &  6.81     & 8.07 \\
    GPT-4o~\cite{openai2025introducing} &  \textbf{8.96}     &  8.91    &       &  \textbf{8.90}     &   \textbf{8.95}    &  \textbf{8.81}     &       &   \textbf{8.92}    &   \textbf{8.40}    &   \textbf{8.44}    & \textbf{8.78} \\
    
    \midrule
    \rowcolor{blue4!20} \multicolumn{12}{c}{\textit{\textbf{Generation model}}} \\
    \midrule
    FLUX.1 Kontext [dev]~\cite{labs2025flux1kontextflowmatching} & 8.07  & 7.97  &       & -     & -     & -     &       & -     & -     & -     & - \\
    UNO~\cite{wu2025less} &  7.15     &   6.72    &       &  3.56     & 6.46      & 4.90      &       &   2.72    &  4.89     &   4.76    & 5.14 \\
    USO~\cite{wu2025uso} &  8.03     &   7.55    &       &  3.32     &    6.10   &    4.56   &       &   2.77    &  5.38     &   5.09    & 5.35 \\
    UniWorld-V2~\cite{li2025uniworld} &  8.45     &   \textbf{8.44}    &       &   7.87    &   8.22    &   \textbf{7.95}    &       &   \textbf{5.36}    &   7.47    &  \textbf{6.98}     & 7.59 \\
    Qwen-Image-Edit-2509~\cite{wu2025qwenimagetechnicalreport} & \textbf{8.56}      & 8.41  &    &  \textbf{7.92}     &  \textbf{8.37}     &  7.79  &   & 5.23      & \textbf{7.70}      &  6.86     &  \textbf{7.60}      \\
    \midrule

    \rowcolor[HTML]{E6E6FA} \multicolumn{12}{c}{\textit{\textbf{Unified model}}} \\
    \midrule
    BAGEL$^{\dagger}$~\cite{deng2025bagel} &  7.00     &  7.04     &       &  5.32     &  6.69     &  6.74     &       & 3.94      & 5.77      & 5.73      & 6.03 \\
    OmniGen2~\cite{wu2025omnigen2} &  8.17     &  7.63  &   &  7.26     &  7.03     &   7.56  &  &  7.02     &  6.90     &  6.64     &  7.28     \\
    Echo-4o~\cite{ye2025echo4oharnessingpowergpt4o} & \textbf{8.34}      &  8.27     &       &  8.13     &  \textbf{8.14}     &   8.11    &       & \textbf{7.07}      & 7.73      &  \textbf{7.77}     & 7.95 \\

    \midrule
    Scone (Ours) &  \textbf{8.34}     &  \textbf{8.52}     &       &  \textbf{8.24}    &   \textbf{8.14}    &   \textbf{8.30}  &  &   7.06    &   \textbf{7.88}    &   7.63    &   \textbf{8.01} \\
    \bottomrule
    \end{tabular}}
  \label{tab:omnicontext}%
\end{table*}%

\begin{table*}[htbp]
  \centering
  \caption{\textbf{Quantitative comparison of existing models on our~\ourbench~benchmark.} ${\dagger}$ indicates our base model. ``COM'': Composition score. ``DIS'': Distinction score.
  Best scores in each group are highlighted in \textbf{bold}.
  }
  
    \resizebox{1\textwidth}{!}{
    \begin{tabular}{lccccccccccccccccccc}
    \toprule
    \multirow{3}[6]{*}{\textbf{Method}} & \multicolumn{3}{c}{\textbf{Composition~$\uparrow$}} &       & \multicolumn{5}{c}{\textbf{Distinction~$\uparrow$}} &       & \multicolumn{5}{c}{\textbf{Distinction \& Composition~$\uparrow$}} &       & \multicolumn{3}{c}{\multirow{2}[4]{*}{\textbf{Average~$\uparrow$}}} \\
\cmidrule{2-4}\cmidrule{6-10}\cmidrule{12-16}          & \textbf{Single} &       & \textbf{Multi} &       & \multicolumn{2}{c}{\textbf{Cross}} &       & \multicolumn{2}{c}{\textbf{Intra}} &       & \multicolumn{2}{c}{\textbf{Cross}} &       & \multicolumn{2}{c}{\textbf{Intra}} &       & \multicolumn{3}{c}{} \\
\cmidrule{2-2}\cmidrule{4-4}\cmidrule{6-7}\cmidrule{9-10}\cmidrule{12-13}\cmidrule{15-16}\cmidrule{18-20}          & \textbf{COM} &       & \textbf{COM} &       & \textbf{COM} & \textbf{DIS} &       & \textbf{COM} & \textbf{DIS} &       & \textbf{COM} & \textbf{DIS} &       & \textbf{COM} & \textbf{DIS} &       & \textbf{COM} & \textbf{DIS} & \textbf{Overall} \\
    \midrule
    \rowcolor{red4!20} \multicolumn{20}{c}{\textit{\textbf{Closed-Source Model}}} \\
    \midrule
    Gemini-2.5-Flash-Image~\cite{google2025gemini} & 8.87  &       & 7.94  &       & 9.12  & \textbf{9.15}  &       & 9.00  & 8.50  &       & 8.27  & \textbf{8.87}  &       & 8.17  & 8.85  &       & 8.56  & 8.84  & 8.70  \\
    GPT-4o~\cite{openai2025introducing} & \textbf{8.92}  &       & \textbf{8.51}  &       & \textbf{9.18}  & 8.55  &       & \textbf{9.45}  & \textbf{9.01}  &       & \textbf{8.83}  & 8.49  &       & \textbf{8.99}  & \textbf{9.56}  &       & \textbf{8.98}  & \textbf{8.90}  & \textbf{8.94}  \\

    \midrule
    \rowcolor{blue4!20} \multicolumn{20}{c}{\textit{\textbf{Generation Model}}} \\
    \midrule
    FLUX.1 Kontext [dev]~\cite{labs2025flux1kontextflowmatching} & 7.92  &       & -     &       & 7.93  & 8.45  &       & 6.20  & 6.11  &       & -     & -     &       & -     & -     &       & -     & -     & - \\
    USO~\cite{wu2025uso}   & 8.03  &       & 5.19  &       & 7.96  & 8.50  &       & 7.14  & 6.51  &       & 5.10  & 6.25  &       & 5.07  & 5.57  &       & 6.41  & 6.71  & 6.56  \\
    UNO~\cite{wu2025less}   & 7.53  &       & 5.38  &       & 7.27  & 7.90  &       & 6.76  & 6.53  &       & 5.27  & 7.02  &       & 5.61  & 6.27  &       & 6.31  & 6.93  & 6.62  \\
    UniWorld-V2~\cite{li2025uniworld} &   8.41  &       & \textbf{7.16}  &       & 8.63  & 8.24  &       & \textbf{7.44}  & 6.77  &       & 7.52  & 8.03  &       & \textbf{7.70}  & \textbf{7.24}  &       & \textbf{7.81}  & 7.57  & 7.69  \\
    Qwen-Image-Edit-2509~\cite{wu2025qwenimagetechnicalreport}  &  \textbf{8.54}  &       & 6.85  &       & \textbf{8.85}  & \textbf{8.57}  &       & 7.32  & \textbf{6.86}  &       & \textbf{7.53}  & \textbf{8.13 } &       & 7.49  & 7.02  &       & 7.76  & \textbf{7.65}  & \textbf{7.70}  \\
    \midrule
    \rowcolor[HTML]{E6E6FA} \multicolumn{20}{c}{\textit{\textbf{Unified Model}}} \\
    \midrule
    BAGEL$^{\dagger}$~\cite{deng2025bagel} & 7.14  &       & 5.55  &       & 7.49  & 7.95  &       & 6.93  & 6.21  &       & 6.44  & 7.38  &       & 6.87  & 7.27  &       & 6.74  & 7.20  & 6.97  \\
    OmniGen2~\cite{wu2025omnigen2} & 8.00  &       & 6.59  &       & 8.31  & 8.99  &       & 6.99  & 6.80  &       & 7.28  & 8.30  &       & 7.14  & 7.13  &       & 7.39  & 7.81  & 7.60  \\
    Echo-4o~\cite{ye2025echo4oharnessingpowergpt4o} & \textbf{8.58}  &       & \textbf{7.73}  &       & 8.36  & 8.33  &       & 7.74  & 7.18  &       & 7.87  & 8.72  &       & 8.01  & 8.33  &       & 8.05  & 8.14  & 8.09  \\
    \midrule
    Scone (Ours)  & 8.52  &       & 7.40  &       & \textbf{8.98}  & \textbf{9.73}  &       & \textbf{7.97}  & \textbf{7.74}  &       & \textbf{8.20}  & \textbf{9.25}  &       & \textbf{8.21}  & \textbf{8.44}  &       & \textbf{8.21}  & \textbf{8.79}  & \textbf{8.50}  \\
    \bottomrule
    \end{tabular}}
  \label{tab:sconebench}%
\end{table*}%

\begin{figure*}[t]
    \centering
    \includegraphics[width=1\textwidth]{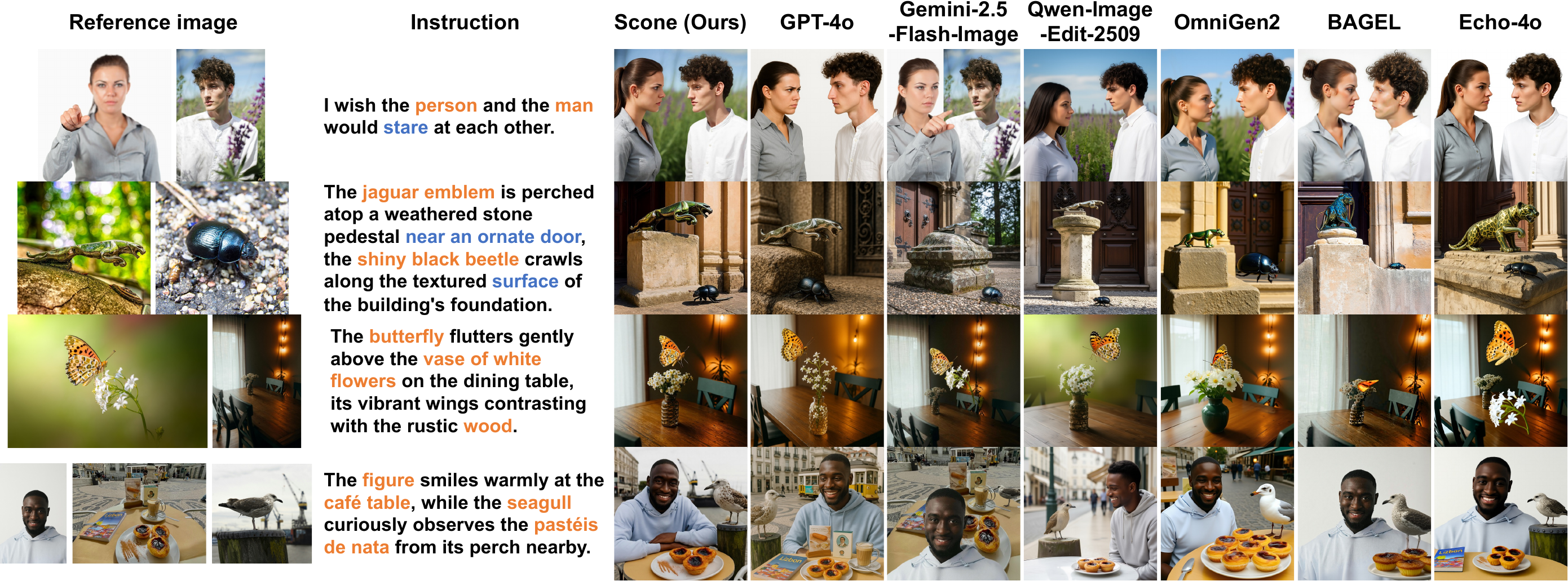}
    \caption{\textbf{Qualitative comparison of existing models on OmniContext~\cite{wu2025omnigen2} benchmark.}}
    \label{fig:qualitative_omnicontext}
\end{figure*}

\section{Experiments}
\subsection{Implementation details}

\paragraph{Training data}
\label{sec:training_data}

We collect a large-scale pool of open-source subject-driven generation datasets, including X2I~\cite{xiao2025omnigen}, MUSAR-Gen~\cite{guo2025musar}, UNO-1M~\cite{wu2025less}, and Echo-4o-Image~\cite{ye2025echo4oharnessingpowergpt4o}.
We further synthesize 15K samples with 3-4 input images to supplement missing data types.
Data categories cover character, object, and scene with predefined object types and attributes.
GPT-4o~\cite{openai2025gpt4o} generates prompts, instructions, and descriptions from random attribute combinations, FLUX.1-dev~\cite{flux2024} produces input images, and Gemini-2.5-Flash-Image~\cite{google2025gemini} generates the final outputs.
\textbf{(1) For training stage \uppercase\expandafter{\romannumeral1}:}
We randomly select \textbf{70K base single-candidate samples} from the data pool and further filter \textbf{22K refined single-candidate samples} using \qwenvl~\cite{qwen3technicalreport} by scoring subject consistency and instruction following.
\textbf{(2) For training stage \uppercase\expandafter{\romannumeral2}:}
Besides the refined single-candidate data, we construct \textbf{20K multi-candidate samples} from another filtered subset. For image acquisition, \qwen~\cite{qwen3technicalreport} extracts subject names and scene descriptions from single-candidate instructions, and \qwenedit~\cite{wu2025qwenimagetechnicalreport} adds cross- or intra-category subjects to create multi-candidate images. The original image serves as the target.
For instruction construction, original instructions are reused for cross-category data. For intra-category data, \qwenvl~\cite{qwen3technicalreport} identifies the new subject introduced during editing, and \qwen~\cite{qwen3technicalreport} replaces the subject name in the original instruction (\eg ``the woman'' $\to$ ``the woman on the left of the image'').

\paragraph{Evaluation settings}
We evaluate on OmniContext~\cite{wu2025omnigen2} and \ourbench. Images are sampled at 1024 $\times$ 1024 with each method’s default settings. To reduce randomness, we conduct 3 sampling rounds and score each round 3 times, yielding 9 result groups whose average is reported.

\subsection{Quantitative and qualitative evaluation}
\paragraph{Quantitative evaluation}
On OmniContext, as shown in~\cref{tab:omnicontext}, our \ourmethod~achieves the highest average score among open-source methods, showing strong composition capability. Closed-source models GPT-4o~\cite{openai2025introducing} and Gemini-2.5-Flash-Image~\cite{google2025gemini} achieve the top two average scores, demonstrating leading performance.
On \ourbench, as shown in~\cref{tab:sconebench}, \ourmethod~achieves the best composition, distinction and overall scores among open-source models, showing strong composition and distinction performance.
Unified models with lower composition scores, such as OmniGen2~\cite{wu2025omnigen2}, still outperform generation models like Qwen-Image-Edit-2509~\cite{wu2025qwenimagetechnicalreport} in distinction, highlighting the benefit of understanding for subject distinction.
GPT-4o~\cite{openai2025introducing} and Gemini-2.5-Flash-Image~\cite{google2025gemini} exhibit strong composition and distinction abilities, securing the top two overall scores, consistent with results on OmniContext.
Moreover, generation in complex contexts remains difficult due to semantic/visual interference and unstable subject preservation. In~\cref{fig:stability}, \ourmethod~achieves the lowest score standard deviation on \ourbench, indicating the best stability.

\paragraph{Qualitative evaluation}

\begin{figure*}[t]
    \centering
    \includegraphics[width=1\textwidth]{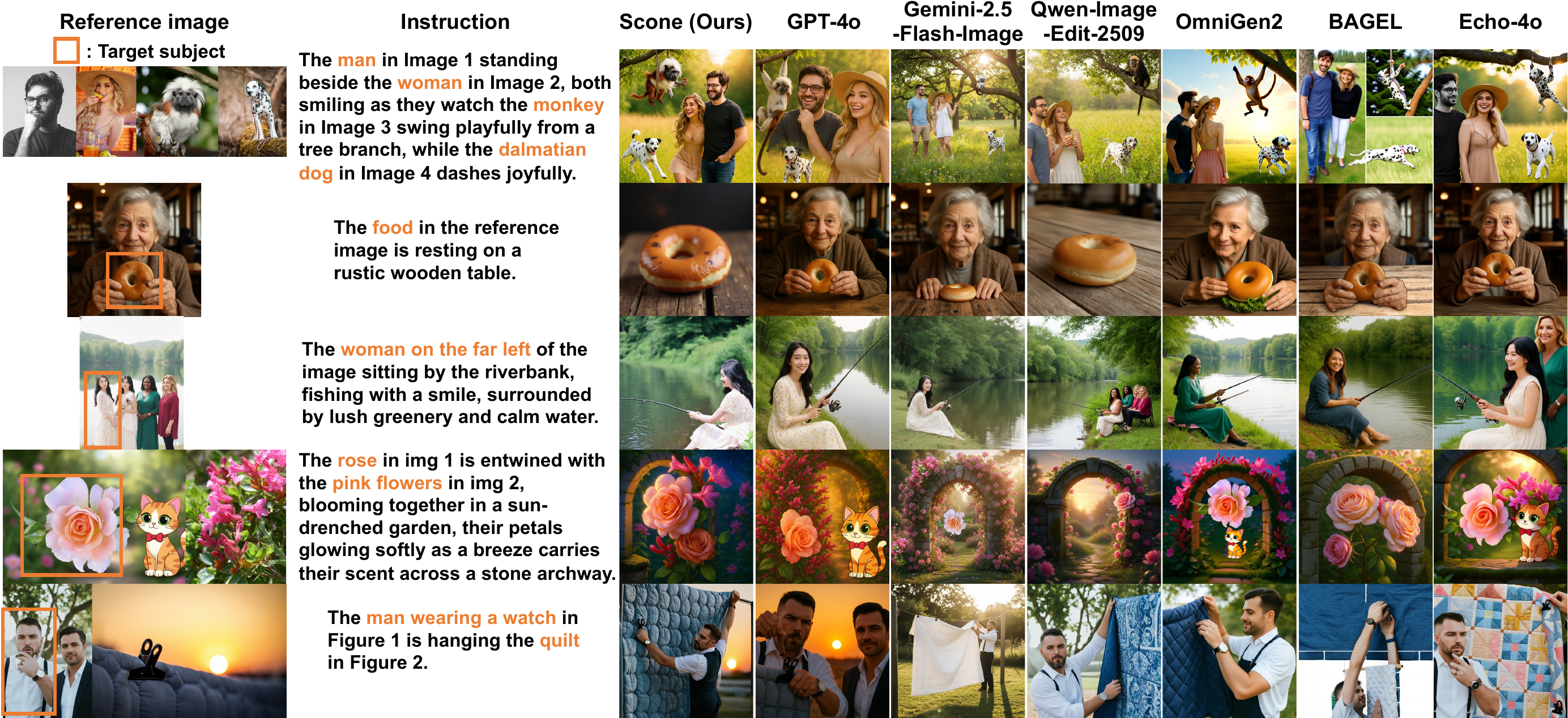}
    \caption{\textbf{Qualitative comparison of existing models on \ourbench~benchmark.}}
    \label{fig:qualitative_sconeeval}
\end{figure*}

\begin{figure}[t]
    \centering
    \includegraphics[width=0.96\columnwidth]{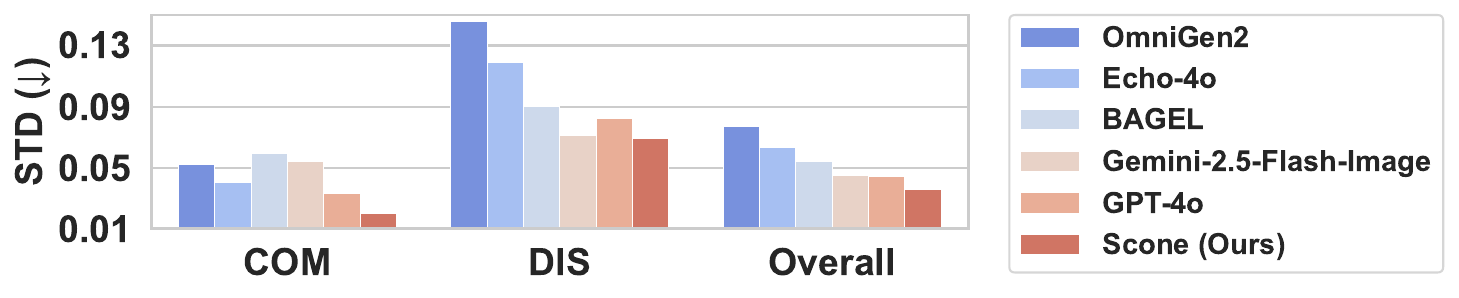}
    \caption{\textbf{Stability measured by score standard deviation.
    }}
    \label{fig:stability}
\end{figure}

Results on OmniContext in~\cref{fig:qualitative_omnicontext} show that \ourmethod~generates natural compositions with strong subject consistency. Results on \ourbench~in~\cref{fig:qualitative_sconeeval} further demonstrate that \ourmethod~can compose four subjects and distinguish the target subject among multiple candidates, reducing issues such as subject redundancy, blending, and omission. All results are sampled with the same seed.

\begin{table}[t]
    \centering
    \caption{\textbf{Ablation results for stage \uppercase\expandafter{\romannumeral1}.} Evaluated on OmniContext benchmark. ``PF'': prompt following. ``SC'': subject consistency.
    }

    \resizebox{1\columnwidth}{!}{
        \begin{tabular}{ccccc}
        \toprule
        \textbf{Version} & \textbf{Training data} & \textbf{PF~$\uparrow$} & \textbf{SC~$\uparrow$} & \textbf{Overall~$\uparrow$} \\
        
        \midrule
        BAGEL~\cite{deng2025bagel} &  - & 6.74 & 5.73 & 6.03 \\
        
        \midrule
        Stage \uppercase\expandafter{\romannumeral1}, step 1     & Base single-candidate data (70K)   &  7.83     &  8.27     & 7.95 \\
        Stage \uppercase\expandafter{\romannumeral1}, step 2     & Refined single-candidate data (22K)   &  \textbf{7.92}     &   \textbf{8.31}    & \textbf{8.02} \\
        \bottomrule
        \end{tabular}}
      \label{tab:abl_data}%
\end{table}%

\begin{table}[tbp]
  \centering
  \caption{\textbf{Ablation results for stage \uppercase\expandafter{\romannumeral2}.} Evaluated on \ourbench~benchmark. ``COM'': composition. ``DIS'': distinction.
  }
  \resizebox{0.9\columnwidth}{!}{
    \begin{tabular}{clccc}
    \toprule
    \textbf{Stage} & \textbf{Version} & \textbf{COM~$\uparrow$} & \textbf{DIS~$\uparrow$} & \textbf{Overall~$\uparrow$} \\
    \midrule
    Base  & BAGEL~\cite{deng2025bagel} & 6.74  & 7.20  & 6.97  \\
    \midrule
    Stage \uppercase\expandafter{\romannumeral1} & -     & 7.94  & 7.78  & 7.86  \\
    \midrule
    \multirow{3}[2]{*}{Stage \uppercase\expandafter{\romannumeral2}} & (a) Direct & 7.64  & 8.23  & 7.94  \\
          & (b) Two-step, w/o bridge & 8.15  & 8.70  & 8.43  \\
          & (c) Two-step, w/ bridge (Ours) & \textbf{8.21} & \textbf{8.79} & \textbf{8.50} \\
    \bottomrule
    \end{tabular}}
  \label{tab:abl_bridge}%
\end{table}%

\begin{table}[tbp]
  \centering
  \caption{\textbf{Parameter study of threshold in stage \uppercase\expandafter{\romannumeral2}.}
  Evaluated on \ourbench~benchmark.
  ``COM'': composition. ``DIS'': distinction.
  }
  \resizebox{0.65\columnwidth}{!}{
    \begin{tabular}{lccc}
    \toprule
          \textbf{Threshold $\boldsymbol{\tau}$} & \textbf{COM~$\uparrow$} & \textbf{DIS~$\uparrow$} & \textbf{Overall~$\uparrow$} \\
  \midrule
    w/o bridge & 8.15 & 8.70 & 8.43 \\
    \midrule
    0.82  & 8.18  &  8.73  &  8.46  \\
    0.85  & 8.19  &  8.75  &  8.47 \\
    0.88  & \textbf{8.21}  & \textbf{8.79}   &  \textbf{8.50}  \\
    
    \bottomrule
    \end{tabular}}
  \label{tab:supp_parameter}%
\end{table}%

\begin{figure}
    \centering
    \includegraphics[width=0.94\columnwidth]{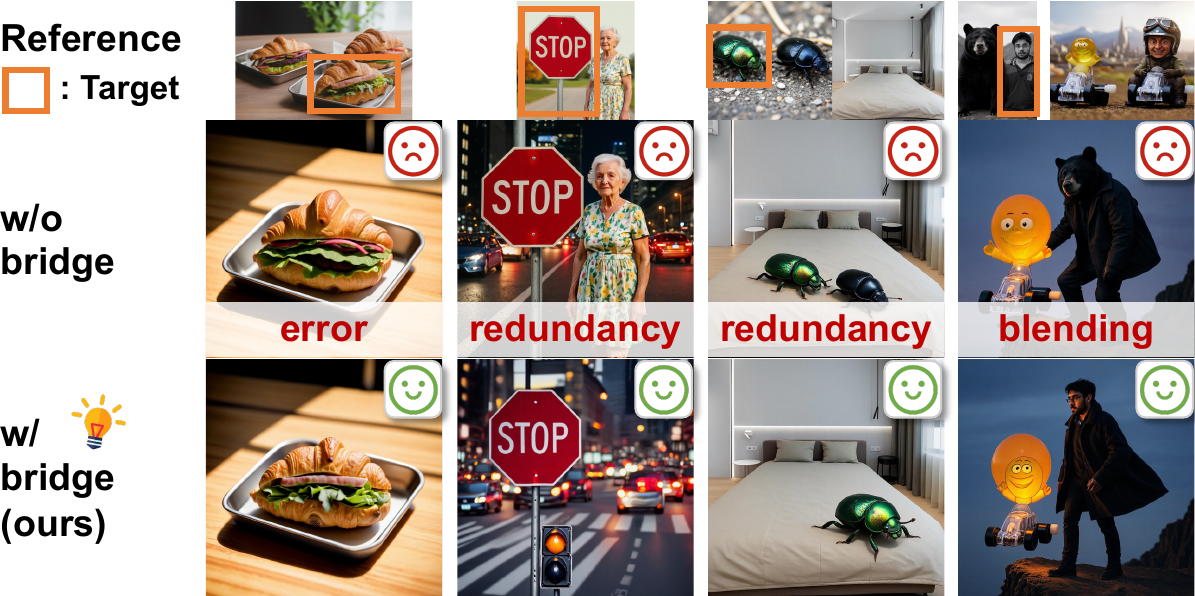}
    \caption{\textbf{Qualitative comparison of bridge on \ourbench.
    }}
    \label{fig:abl_bridge}
\end{figure}

\subsection{User study}
\label{sec:user_study}
We validate the alignment of GPT-4.1 scores with human evaluation. Thirty evaluators, both professionals and non-professionals, assess cases from \ourbench. Each evaluator reviews 60 test samples, with 10 samples from each subtask, and compares the outputs of OmniGen2~\cite{wu2025omnigen2}, UniWorld-V2~\cite{li2025uniworld} and our \ourmethod~side by side. Evaluators select the best result based on instruction following, subject consistency, realism, and aesthetics. After normalization, the scores are: OmniGen2 0.27, UniWorld-V2 0.27, and \ourmethod~0.46, confirming both the reasonableness of GPT-4.1 scores and the effectiveness of our method.

\subsection{Ablation study and parameter study}
\paragraph{Effects of refined data in stage \uppercase\expandafter{\romannumeral1}}

As shown in~\cref{tab:abl_data}, stage \uppercase\expandafter{\romannumeral1} significantly improves composition performance over BAGEL. The 70K base set brings major gains, and the refined 22K set further boosts both prompt following and subject consistency, highlighting that role of data quality.

\paragraph{Effects of understanding bridge strategy in stage \uppercase\expandafter{\romannumeral2}}
We compare three versions, (a) direct fine-tuning of both the understanding and generation experts, and (b)(c) first fine-tuning the understanding expert and then both experts, differing in bridge usage. All are trained for 2k steps, with 1k per stage in the two-step setting.
As shown in~\cref{tab:abl_bridge}, the two-step strategy outperforms direct fine-tuning, and the bridge further improves subject distinction and overall robustness.
Qualitative results in~\cref{fig:abl_bridge}, spanning cross- and intra-category cases from the Distinction and Composition \& Distinction tasks, further highlight the bridge’s benefit for more robust distinction.
An additional user study, following~\cref{sec:user_study}, compares Echo-4o, Scone w/o bridge, and Scone w/ bridge, yielding scores of 0.27, 0.31, and 0.42, respectively, again showing the bridge version performs best.

\paragraph{Parameter study of threshold in stage \uppercase\expandafter{\romannumeral2}}
\label{sec:param}
As shown in~\cref{tab:supp_parameter}, reducing irrelevant token interference steadily improves performance, showing robust semantic guidance.

\section{Conclusion}
We introduce \textbf{\ourmethod}, a unified understanding-generation framework that addresses a neglected problem in subject-driven generation: distinguishing target subjects in multi-candidate contexts. By using the understanding expert as a semantic bridge, it aligns semantics early and filters irrelevant content, guiding the generation expert toward accurate subject preservation and robust composition.
Together with \textbf{\ourbench}, it offers a comprehensive benchmark for evaluating and improving both composition and distinction.

\clearpage
\section*{Acknowledgments}
This work is supported by National Natural Science Foundation of China (92470121, 62402016), Fundamental and Interdisciplinary Disciplines Breakthrough Plan of the Ministry of Education of China (JYB2025XDXM113), National Key R\&D Program of China (2024YFA1014003), Zhongguancun Academy (C20250204, C20250602),  Beijing Major Science and Technology Project (Z251100008125043, Z251100008425023), and High-performance Computing Platform of Peking University.

{
    \small
    \bibliographystyle{ieeenat_fullname}
    \bibliography{main}
}

\clearpage
\maketitlesupplementary
\appendix

\section{Additional details of motivation}
\label{sec:supp_motivation}
The similarity visualizations of instruction and image tokens in Figs.1(b) and 2(a) of the main paper are based on our base model, BAGEL~\cite{deng2025bagel}.
To better highlight high-similarity regions, we retain the top 50\% and generate masked images.
We group layers into four sets (0-7, 8-15, 16-23, 24-27) based on the layer function analysis from~\cite{Zhang_2025_CVPR}. In~\cref{fig:supp_motivation}, we further show representative similarity and masked images for each group.

\paragraph{Observation 1 (Comparison between understanding and generation experts)}
The understanding expert captures more distinct semantic information from the image than the generation expert, as its image-token hidden states exhibit higher similarity to the instruction-token hidden states. It attends more strongly to instruction-relevant regions, such as candidate subjects.

\paragraph{Observation 2 (Comparison across layers of the understanding expert)}
Although similarity remains high at layers 16 and 24, region discrimination is strongest at layer 8, which provides more distinctive semantic cues for generation guidance. We therefore choose layer 8 to provide the semantic mask, and apply it to the later semantically discriminative layers 9–15.

\section{Additional details of training data}
\subsection{Synthesized data for data pool}
As described in Sec.5.1 of the main paper, we synthesize 15K samples with 3-4 input images to fill gaps in the data pool and improve the composition capability of \ourmethod. Examples are shown in~\cref{fig:supp_training_data_synthesized_3img,fig:supp_training_data_synthesized_4img}.

\begin{figure}[t]
    \centering
    \includegraphics[width=1\columnwidth]{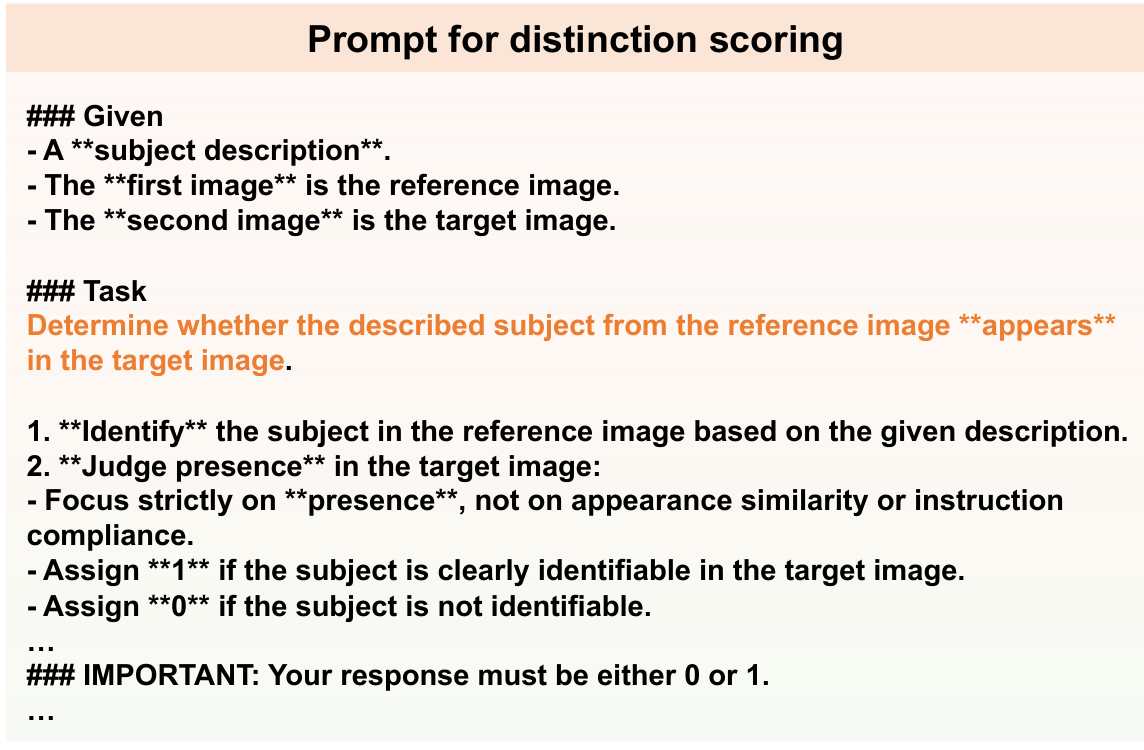}
    \caption{\textbf{Prompt for distinction scoring in \ourbench.} It determines whether the described reference subject \emph{appears} in the target image.}
    \label{fig:supp_benchmark_evaluation_prompt}
\end{figure}

\begin{figure*}[h]
\centering
\includegraphics[width=\textwidth]{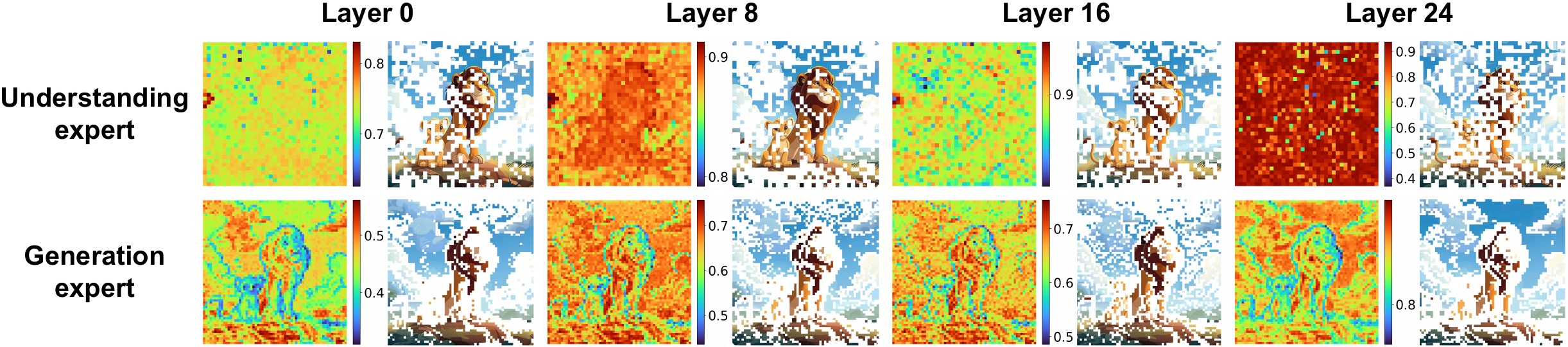}
\caption{\textbf{Representative similarity and masked images for each layer group.} Similarity visualizations of instruction and image token hidden states from understanding and generation experts are based on our base model, BAGEL~\cite{deng2025bagel}. Masked images retain the top 50\% of regions for clearer observation.}
\label{fig:supp_motivation}
\end{figure*}

\subsection{Data filtering for refined single-candidate data}

As described in Sec.5.1 of the main paper, refined single-candidate samples are filtered by scoring subject consistency and instruction following with the VLM model \qwenvl. Key prompt contents are shown in~\cref{fig:supp_training_data_filtering}(a), with emphasis on facial identity, text, and quantity. Each sample is scored from 0 to 4, and only those with a score of 4 are retained, as shown in~\cref{fig:supp_training_data_filtering}(b).

\subsection{Details of multi-candidate data}
\paragraph{Single-subject data}
Multi-candidate single-subject data are derived from single-candidate multi-subject data by reversing the reference and target images, so that the original reference becomes the target and the original target becomes the reference. This avoids the cost of generating new images. For instruction construction, \qwen~\cite{qwen3technicalreport} identifies subjects, provides distinctive descriptions, and generates instructions based on the prompt in~\cref{fig:supp_training_data_multi_candidate_single_subject}(a). The final dataset contains 2 case types, each with cross-category and intra-category candidate subjects. Examples are shown in~\cref{fig:supp_training_data_multi_candidate_single_subject}(b).

\paragraph{Multi-subject data}

Multi-subject data are constructed from single-candidate multi-subject data by editing a subset of the reference images. Specifically, we use GPT-4o~\cite{openai2025gpt4o} to generate prompts for subjects from different categories, and then add at least one subject to the reference images using \qwenedit~\cite{wu2025qwenimagetechnicalreport}.
The instruction construction consists of two steps: \textit{(1) subject identification} and \textit{(2) subject replacement}. Subject identification follows the procedure described in Sec.4.2 of the main paper, with additional details provided in~\cref{sec:supp_benchmark_instruction}.
Subject replacement uses \qwen~\cite{qwen3technicalreport} and the prompt in~\cref{fig:supp_training_data_multi_candidate_multi_subject}(a) to replace the original subject description with the distinct description obtained in Step 1.
The final dataset contains 5 case types, each including both cross-category and intra-category candidate subjects in the reference images. Examples are shown in~\cref{fig:supp_training_data_multi_candidate_multi_subject}(b).

\section{Two-step decoupling instruction construction in \ourbench}
\label{sec:supp_benchmark_instruction}
As described in Sec.4.2 of the main paper, we adopt a two-step decoupling strategy that separates visual understanding from instruction generation, improving instruction stability and quality. As shown in~\cref{fig:supp_benchmark_instruction_comparison}, direct instruction construction often produces unusable results, such as incorrect image indices, ambiguous target subjects, and unrelated subjects.
Our strategy first uses the vision-language model \qwenvl~\cite{qwen3technicalreport} to identify the target subject and generate a distinct description from the raw single-candidate and edited multi-candidate reference images, with prompt in~\cref{fig:supp_benchmark_instruction_prompt}(a). It then uses the language model \qwen~\cite{qwen3technicalreport} to generate instructions solely from the subject descriptions, with prompt in~\cref{fig:supp_benchmark_instruction_prompt}(b).

\begin{figure}[b]
    \centering
    \includegraphics[width=0.7\columnwidth]{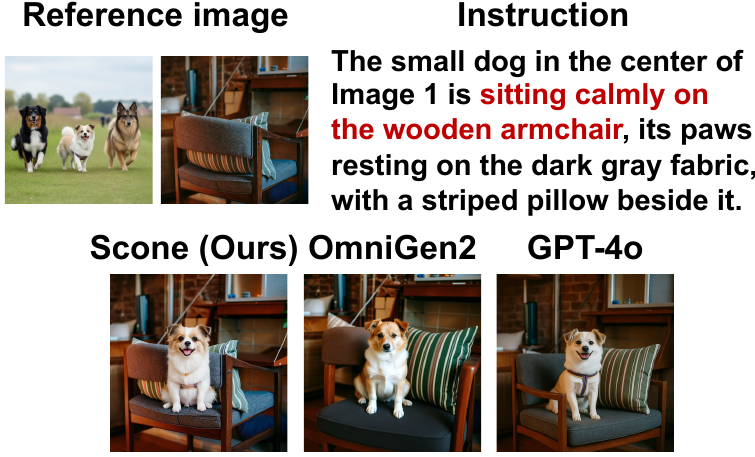}
    \caption{\textbf{Limitation of our~\ourmethod.}}
    \label{fig:supp_limitations}
\end{figure}

\section{Limitation and future work}
Our~\ourmethod~still shares a common limitation with existing methods: unrealistic interactions. As shown in~\cref{fig:supp_limitations}, the generated dog passes through the chair, violating physical laws. This issue appears in both our~\ourmethod~and OmniGen2~\cite{wu2025omnigen2}.
Future work will also explore more efficient mechanisms to reduce redundant image tokens for scalable generation in complex scenarios.

\begin{figure*}[ht]
    \centering
    \includegraphics[width=0.65\textwidth]{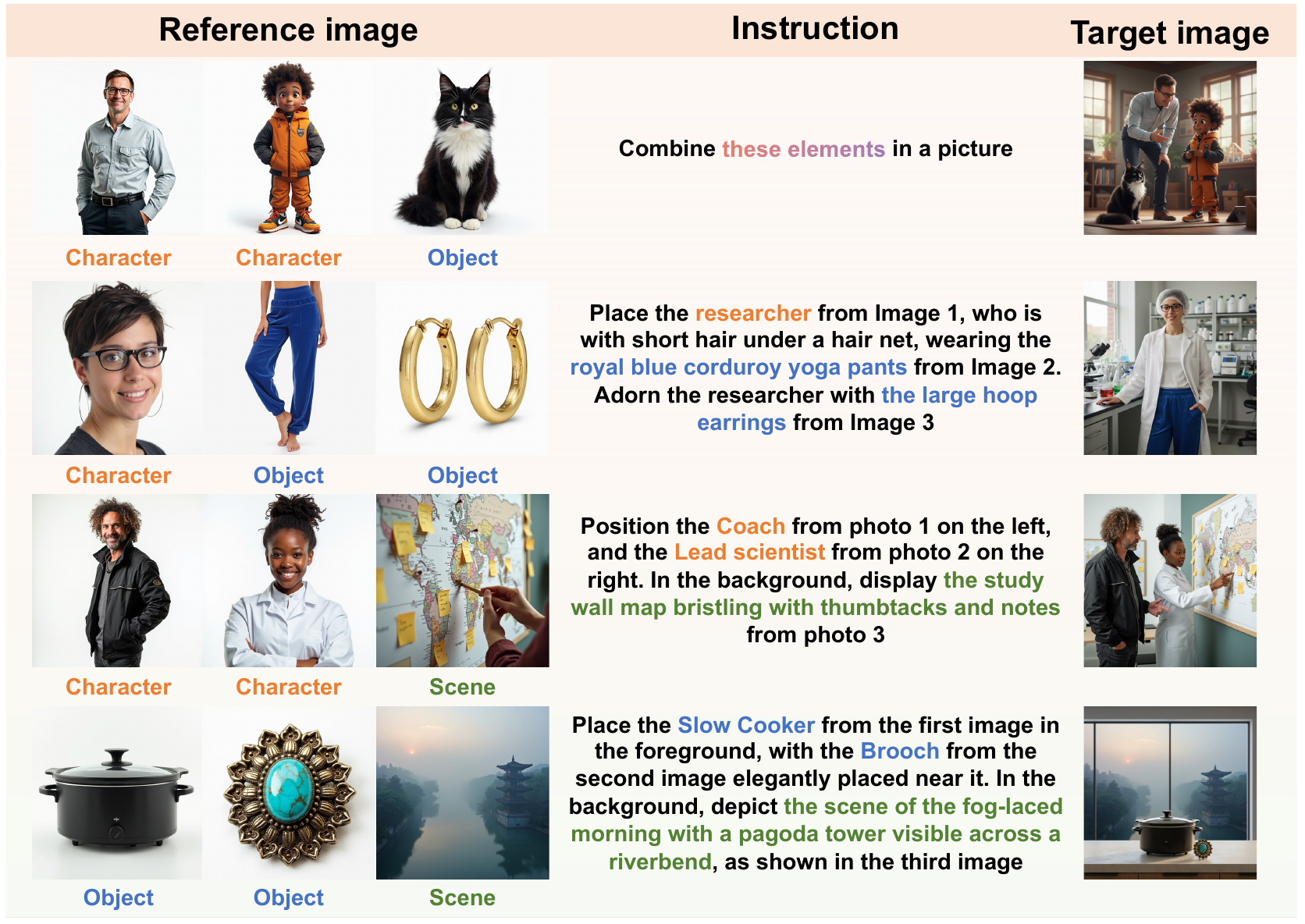}
    \caption{\textbf{Examples of synthesized data with 3 input images.} These cover 4 case types, including character-character/object interactions, characters with multiple objects, characters in scenes, and objects placed in scenes.}
    \label{fig:supp_training_data_synthesized_3img}
\end{figure*}

\begin{figure*}[ht]
    \centering
    \includegraphics[width=\textwidth]{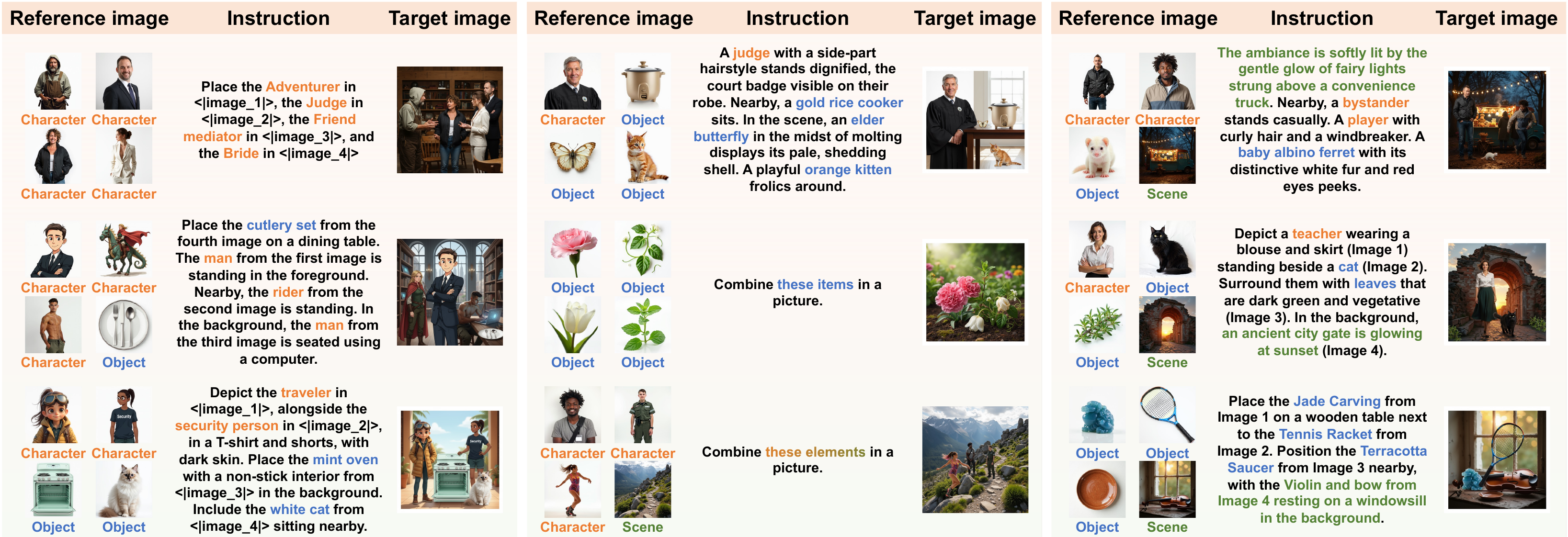}
    \caption{\textbf{Examples of synthesized data with 4 input images.} These cover 9 case types, including combinations of characters, objects, and scenes, as well as their interactions and mixed compositions.}
    \label{fig:supp_training_data_synthesized_4img}
\end{figure*}

\begin{figure*}[ht]
    \centering
    \includegraphics[width=\textwidth]{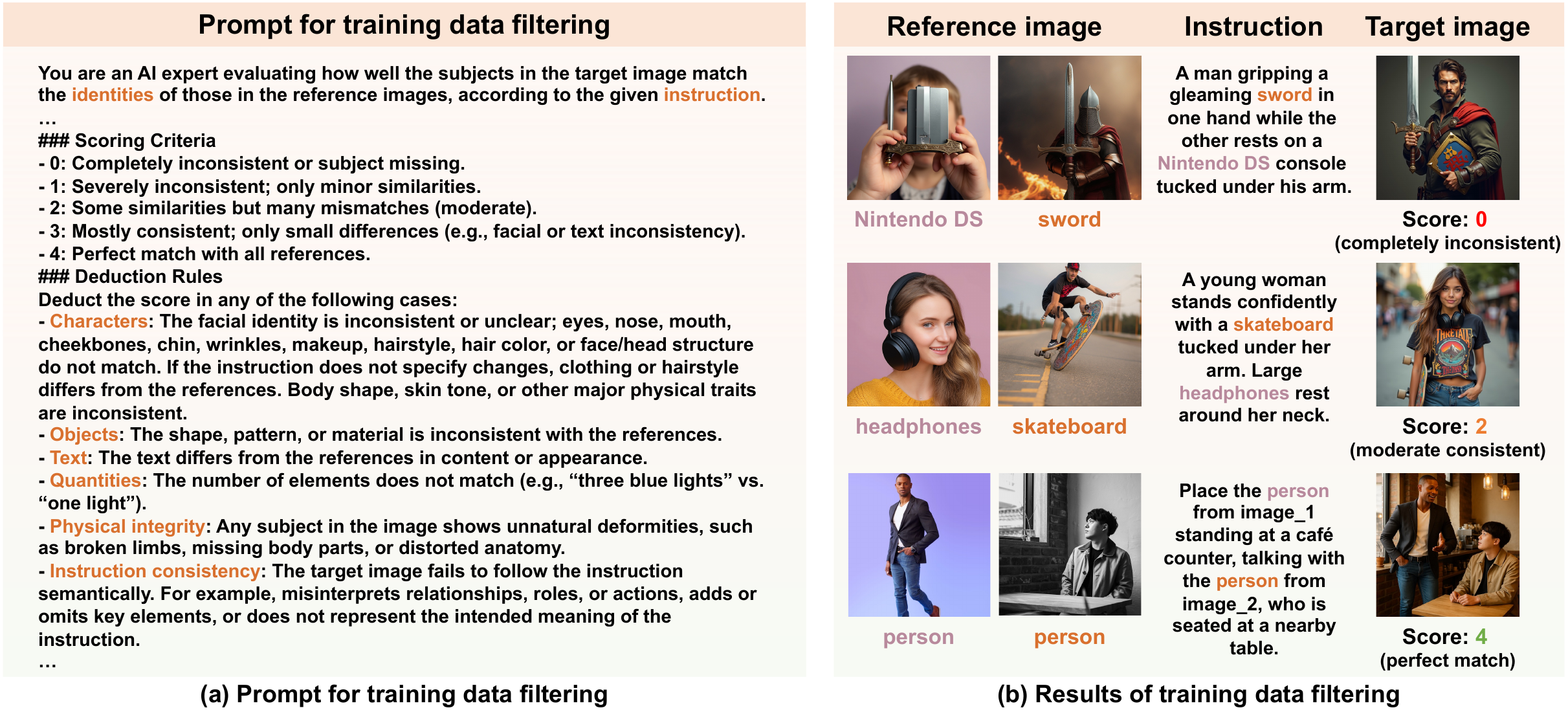}
    \caption{\textbf{Data filtering for refined single-candidate data.} \textbf{(a) Prompt for filtering.} Key prompt components are shown. \textbf{(b) Filtering results.} Samples are scored from 0 to 4, and only those with a score of 4 are retained.}
    \label{fig:supp_training_data_filtering}
\end{figure*}

\begin{figure*}[ht]
    \centering
    \includegraphics[width=0.8\textwidth]{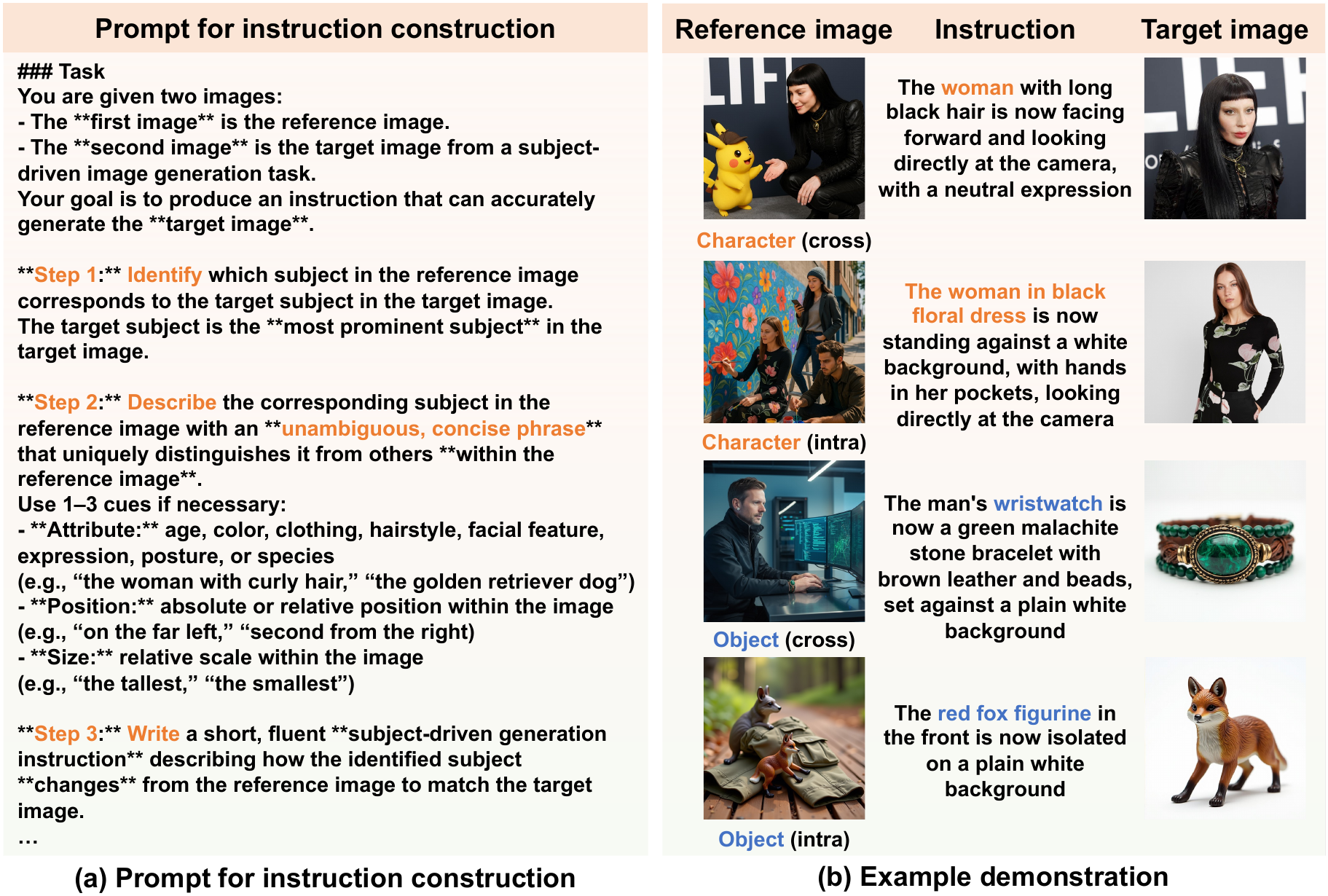}
    \caption{\textbf{Multi-candidate single-subject data construction.}
    \textbf{(a) Prompt for instruction construction.} The prompt instructs the vision-language model to identify subjects, provide distinct descriptions, and generate instructions.
    \textbf{(b) Example demonstration.} It includes 2 case types, Character and Object, each with cross-category and intra-category candidate subjects in the reference images.}
    \label{fig:supp_training_data_multi_candidate_single_subject}
\end{figure*}

\begin{figure*}[ht]
    \centering
    \includegraphics[width=\textwidth]{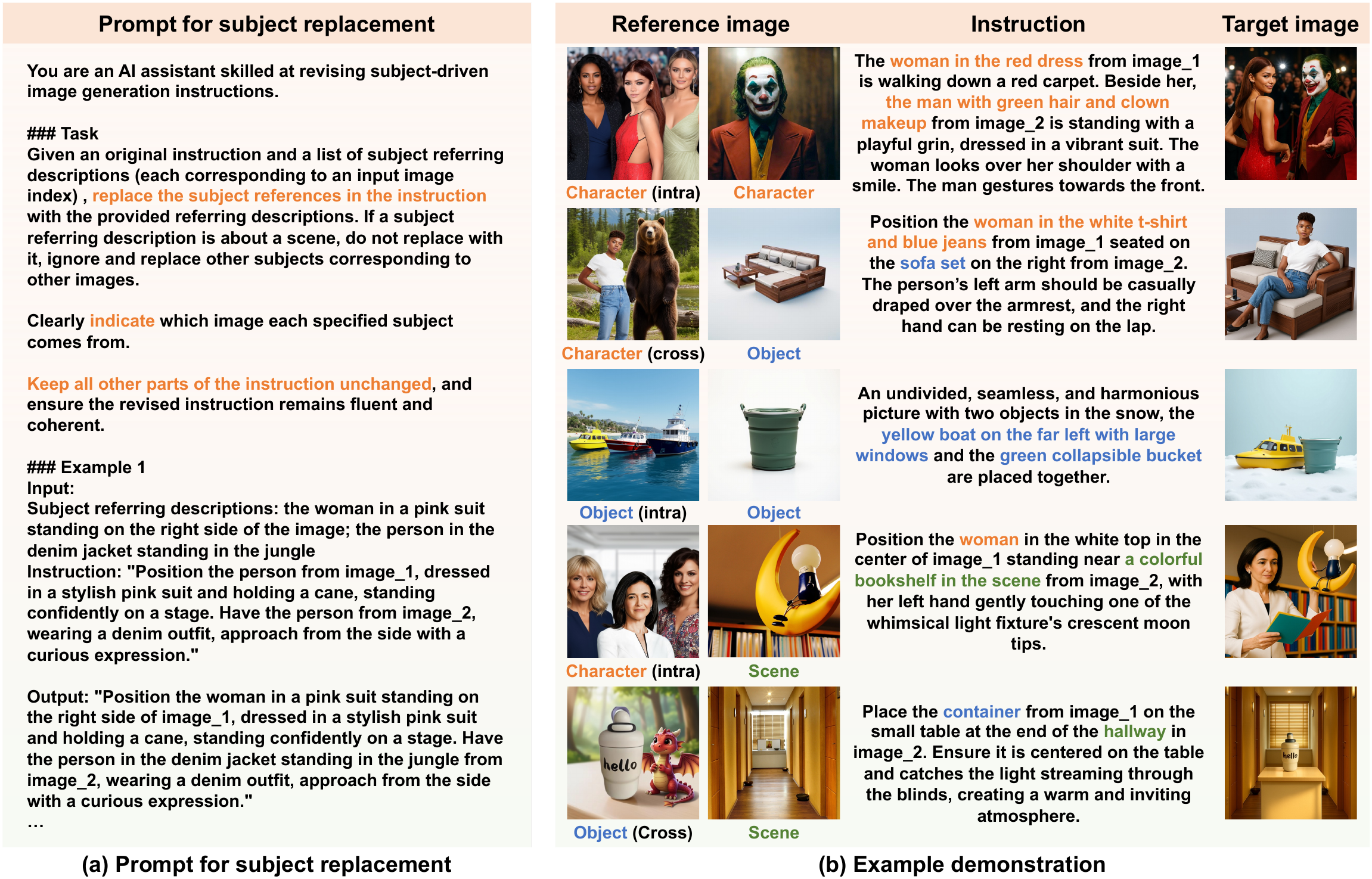}
    \caption{\textbf{Multi-candidate multi-subject data construction.}
    \textbf{(a) Prompts for subject replacement.} The prompts guide the language model to replace the original subject description with a new distinct description for the edited multi-candidate reference images.
    \textbf{(b) Example demonstration.} It includes 5 case types: Character+Character, Character+Object, Object+Object, Character+Scene, and Object+Scene, each with cross-category and intra-category candidate subjects in the reference images.}
    \label{fig:supp_training_data_multi_candidate_multi_subject}
\end{figure*}

\begin{figure*}[ht]
    \centering
    \includegraphics[width=0.6\textwidth]{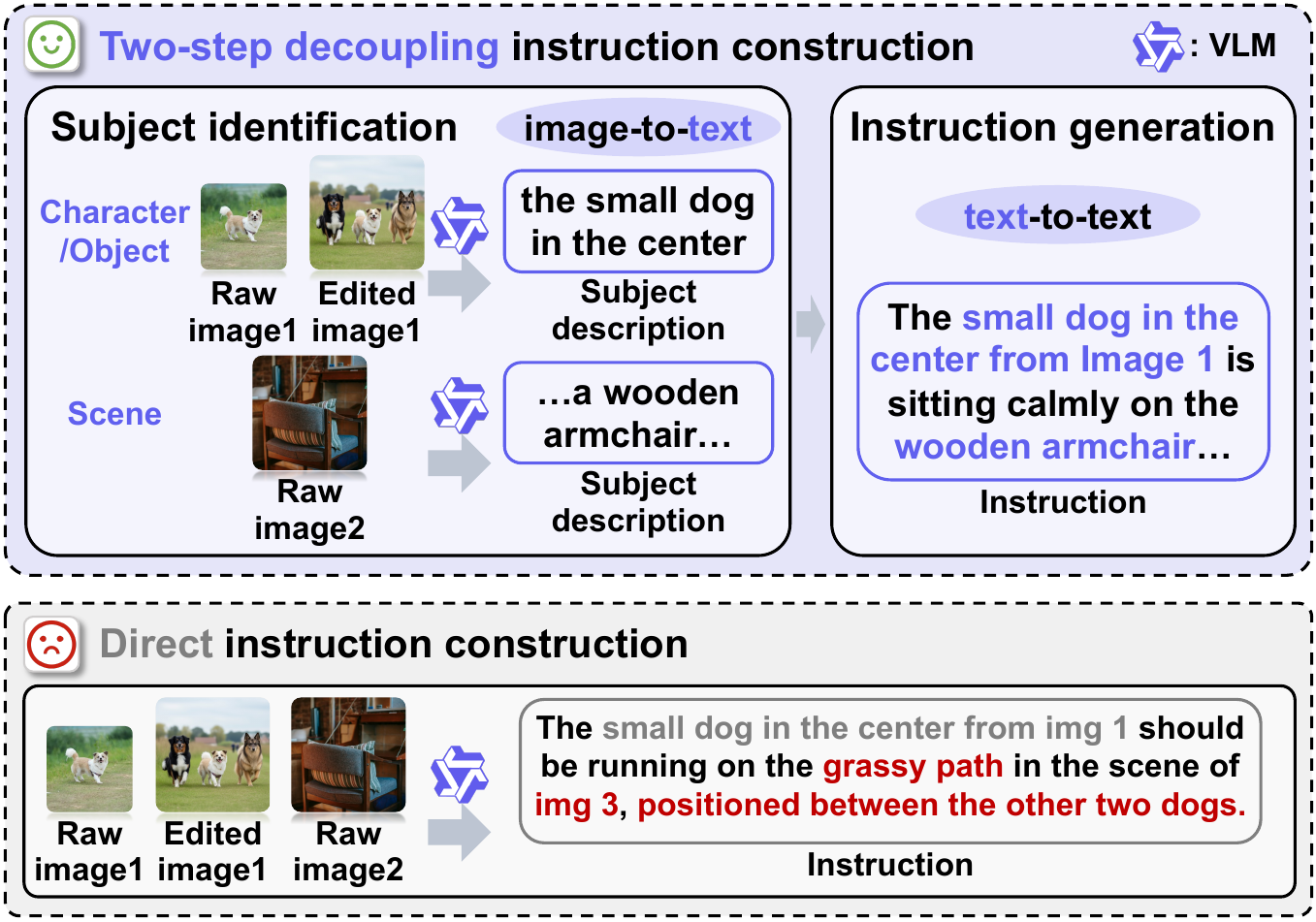}
    \caption{\textbf{Comparison of two instruction construction strategies.} The two-step decoupling strategy separates image-to-text and text-to-text generation, reducing cross-image interference and avoiding errors in the direct strategy, such as incorrect image indices, ambiguous target subjects, and unrelated subjects.}
    \label{fig:supp_benchmark_instruction_comparison}
\end{figure*}

\begin{figure*}[ht]
    \centering
    \includegraphics[width=\textwidth]{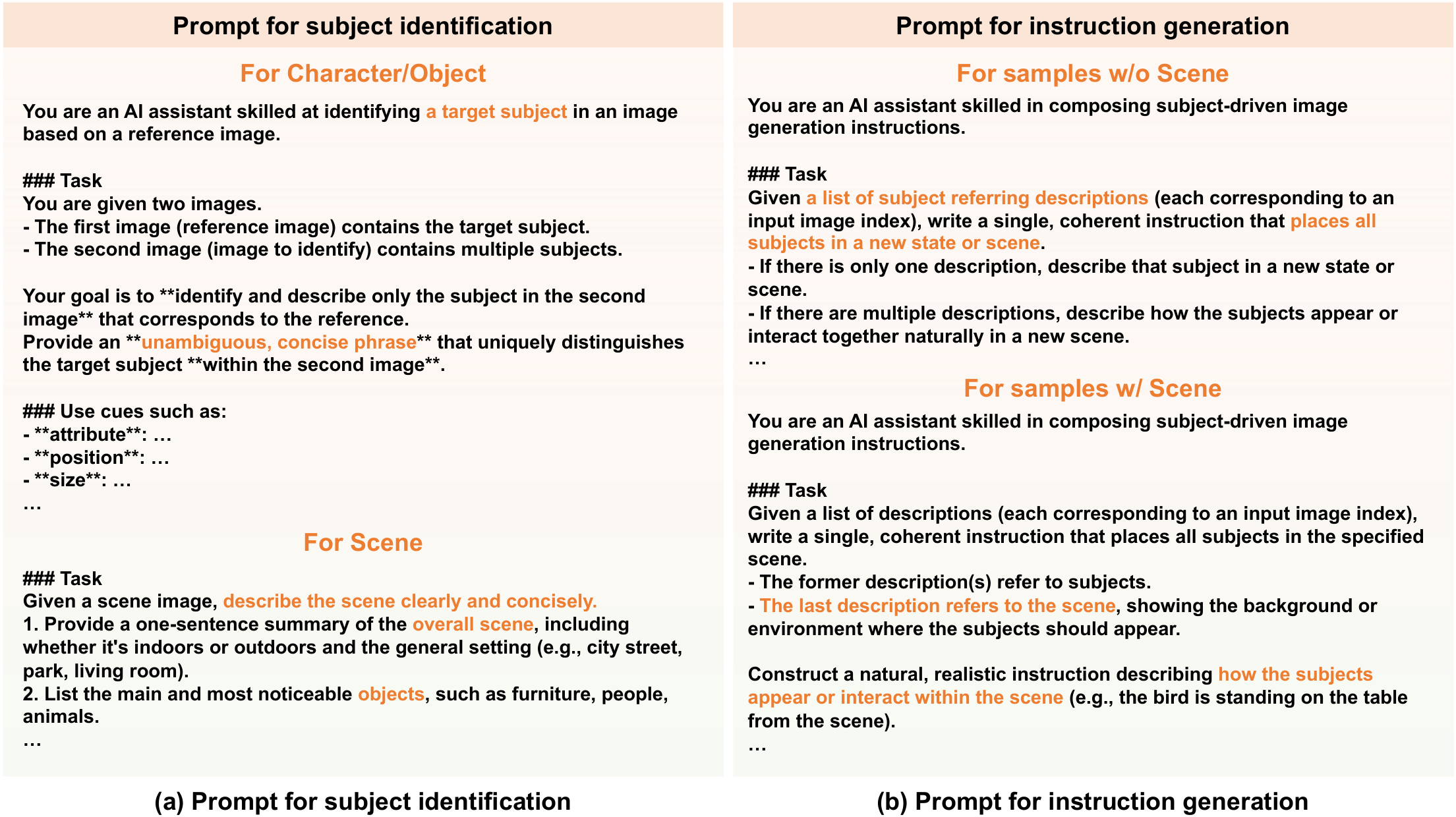}
    \caption{\textbf{Prompts for instruction construction in \ourbench.}
\textbf{(a) Prompt for subject identification.} For Character or Object images, provide a concise subject description; for Scene images, describe the setting and key objects.
\textbf{(b) Prompt for instruction generation.} Generate instructions from the subject descriptions, emphasizing subject-subject and subject-scene interactions.}
    \label{fig:supp_benchmark_instruction_prompt}
\end{figure*}

\end{document}